\pdfoutput=1

\documentclass[11pt]{article}

\usepackage[]{EMNLP2022}

\usepackage{times}
\usepackage{layouts}
\usepackage{graphicx}
\usepackage{latexsym}
\usepackage{amsmath}
\usepackage{theorem}
\usepackage{soul}
\usepackage{multirow}
\usepackage{booktabs}
\usepackage{cuted}
\usepackage{amssymb}
\usepackage{amsfonts}
\usepackage{subfig}

\newtheorem{theorem}{Theorem}

\newcommand{\modelname}{\textsc{Care}~}
\definecolor{blueviolet}{RGB}{48,84,151}
\definecolor{darkgreen}{RGB}{83,129,53}
\definecolor{reallypink}{RGB}{255,98,176}

\usepackage[T1]{fontenc}

\usepackage[utf8]{inputenc}

\usepackage{microtype}

\usepackage{inconsolata}

\title{Evade the Trap of Mediocrity: Promoting Diversity and Novelty in \\Text Generation via Concentrating Attention}

\author{
    Wenhao Li$^{1,2,3}$,\,  Xiaoyuan Yi$^{5}$,\,   Jinyi Hu$^{1,2,3}$,\,   Maosong Sun$^{1,2,3,4}$\thanks{\ \ Corresponding author. Email: sms@tsinghua.edu.cn}, \, Xing Xie$^{5}$ \\
    $^1$ Department of Computer Science and Technology, Tsinghua University, Beijing \\
    $^2$ Beijing National Research Center for Information Science and Technology \\
    $^3$ Institute for Artificial Intelligence, Tsinghua University, Beijing \\
    $^4$ Jiangsu Collaborative Innovation Center for Language Ability, Jiangsu Normal University, Xuzhou \\
    $^5$ Microsoft Research Asia \\
    \texttt{wh-li20@mails.tsinghua.edu.cn,}\,\texttt{xiaoyuanyi@microsoft.com}
}
\begin{document}
\maketitle
\begin{abstract}
Recently, powerful Transformer architectures have proven superior in generating high-quality sentences. Nevertheless, these models tend to produce dull high-frequency phrases, severely hurting the diversity and novelty of generated text. In this work, we dig into the intrinsic mechanism of this problem and found that sparser attention values in Transformer could improve diversity. To understand such a phenomenon, we first conduct both empirical and theoretical analysis and then attribute it to representation degeneration caused by the attentive mixture of the hidden states during training. We term this process the \emph{Trap of Mediocrity}. To escape from such a trap, we introduce a novel attention regularization loss to control the sharpness of the attention distribution, which is transparent to model structures and can be easily implemented within 20 lines of python code. We prove that this method could be mathematically regarded as learning a Bayesian approximation of posterior attention. Experiments show that our method improved the diversity and novelty of the generated text while maintaining comparable quality on a variety of conditional and unconditional generation tasks.
\end{abstract}

\section{Introduction}
\label{sec:intro}
Natural Language Generation (NLG) plays a crucial role in modern natural language Processing (NLP). With considerable benefits for a wide array of applications such as automatic literary writing~\citep{yi-etal-2018-automatic, pens}, summarization~\citep{HEPOS} and paraphasing~\citep{docupara}, this area has risen to prominence.

Recently, neural architectures based on the prevalent Tranformer~\citep{vaswani2017attention}, \textit{e.g.}, GPT-2 \citep{radford2019language}, have proven highly successful in producing fluent text indistinguishable from the human-written one. Nevertheless, these models tend to remember and generate \emph{dull} and \emph{generic} (usually high-frequency) contents~\citep{curiousdege}, even with highly-diverse ground-truth as training targets and distinct prompts as testing inputs. Such a conundrum, known as one kind of \emph{degeneration problem}~\citep{DBLP:conf/acl/ChoiBGHBF18} in NLG, could cause a poor user experience, especially for genres with high requirements for diversity and novelty, like story and headline generation.
\begin{figure}[t]
\center
\includegraphics[width=0.45 \textwidth]{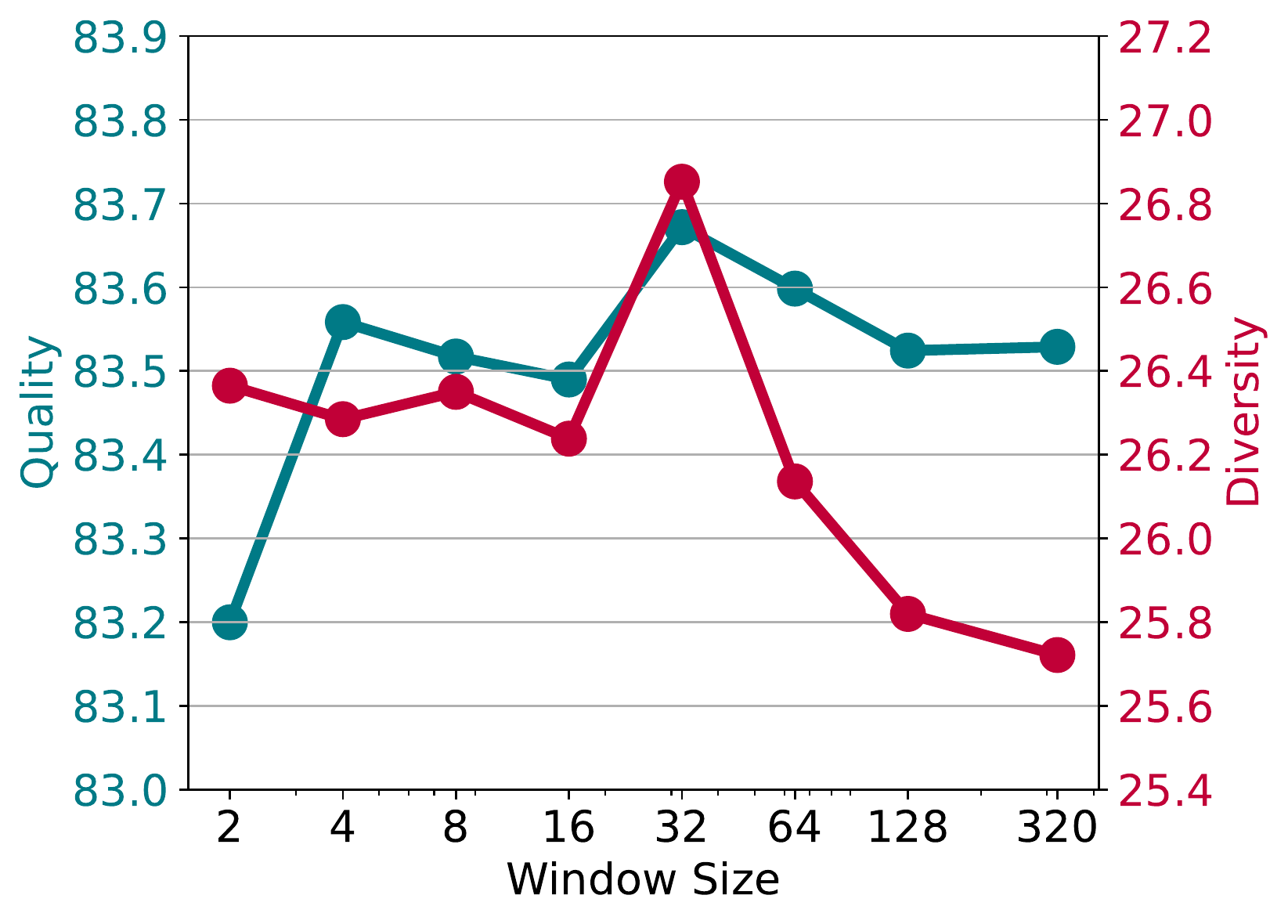}
\caption{Quality and diversity of generated text with varying attention sparsity on ROCStories. The quality and diversity (the higher the better) are measured by BERTScore~\citep{bertscore} and dist~\citep{li-etal-2016-diversity}, respectively. Sparsity is controlled by the attention window size. A smaller window leads to higher sparsity.} 
\label{fig_intro_fig} 
\end{figure}
Several research lines make continued endeavors for this challenge, for instance, adopting stochastic decoding algorithms \citep{DBLP:conf/icml/KoolHW19,curiousdege} or replacing the Maximum Likelihood Estimation (MLE) loss with more sophisticated optimization objectives \citep{DBLP:conf/iclr/WelleckKRDCW20,xu2022learning} to alleviate degeneration. However, these methods are just stopgaps and do not address this issue in depth. The stochastic decoding methods permit the model to choose low-confidence candidates, therefore bringing the contradiction between the quality and the diversity of the generated text. Meanwhile, the methods in the second research line just focus on solving \emph{intra-instance repetition} instead of \emph{inter-instance diversity}. Thus, to intrinsically improve the diversity in text generation without hurting the quality, the underlying mechanism of this problem must be figured out.

\emph{What leads to poor generation diversity and how could we accordingly tackle it?} In this work, we dig into these questions and find another possible cause of such a conundrum in Transformer-based models, namely \emph{Attention Concentration}. Initially inspired by related literary research which demonstrated that the concentration degree of attention on linguistic sources could impact creative writing~\citep{rosenblatt1988writing}, we conducted a preliminary study on the correlation between attention concentration (or sparsity) and generation diversity. We trained a GPT-2 base model with local attention window as in~\citet{sparseTrans} on the ROCStories dataset~\citep{mostafazadeh-etal-2016-roc}. As shown in Fig.~\ref{fig_intro_fig}, obviously, adequate sparsity improves both quality and diversity, while excessive sparsity harms quality. Such improvement of diversity is non-trivial as demonstrated in Sec.~\ref{sec:experiment}.

To better understand such results, we resort to the representation degeneration problem~\citep{RepDeg} and analyze the interaction between the hidden states and word embeddings in Transformer. We discovered that the attentive mixture could impel hidden states to approach the embeddings of high-frequency words, especially when the attention is highly distributed, which encourages irrationally higher probabilities of the generic high-frequency words, harming diversity and novelty. We termed this phenomenon as the \textit{Trap of Mediocrity}.

To escape this trap enforced by evenly dispersed attention, we propose a simple yet effective regularization method, \modelname \footnote{\textbf{C}oncentrating \textbf{A}ttention by \textbf{R}\'{e}nyi \textbf{E}ntropy }, to concentrate and sharpen attention. \modelname reduces the R\'{e}nyi entropy of attention distributions during the training process to enhance sparsity, which is lightweight, transparent to the model architectures, and could be easily implemented within 20 lines of code. We mathematically demonstrate \modelname as learning a Bayesian approximation of the posterior attention given the ground truth, providing a theoretical validation of its superiority. Besides, We also equip \modelname with learnable attention dropout to further boost performance. In this way, \modelname could learn more concentrated attention with minimal computation cost to break the Trap of Mediocrity and hence improve the diversity and novelty of generated text.

In summary, our contributions are as follows:
\begin{itemize}
\item We are the first work to find the correlation between attention concentration and generation diversity, \textit{i.e.}, \emph{Trap of Medicority}, and provide an in-depth analysis of its underlying cause.

\item We propose a lightweight method, \modelname, to escape from such a trap, which is transparent to model structures and easy to implement.

\item We theoretically demonstrate that \modelname can concentrate attention and learn a Bayesian approximation of the posterior attention.

\item Experiments on three conditional and one unconditional generation tasks show that our model could significantly improve generation diversity and novelty against several strong baselines while keeping comparable quality.
\end{itemize}
\section{The Trap of Mediocrity}
\label{sec:analysis}
As introduced in Sec.~\ref{sec:intro}, we find one underlying cause of poor generation diversity lies in the improperly attentive mixture of the hidden states in Transformer, \textit{i.e.}, \emph{Trap of Mediocrity}. To reveal the mechanism of this trap, we performed in-depth analyses both empirically and theoretically. In detail, we progressed our argumentation by proving the following three \textbf{Research Claims (RC)} step by step:

\paragraph{RC1: Formation of the trap.}The output-layer embeddings of high-frequency tokens are more likely to cluster in a uniformly positive direction of corresponding hidden states in the Transformer.

\paragraph{RC2: Falling into the trap} The attentive mixture process of contextual hidden states forces the representations of all tokens to approach that direction, resulting in consistently higher generation probabilities of frequent and mediocre tokens.

\paragraph{RC3: Dispersion dominates the mixture} The extremely dispersed attention distribution (low sparsity) loses the concentration on informative context and dominates the improper mixture both in the formation and falling process.

To support each of these claims, we reuse the GPT-2 in Sec.~\ref{sec:intro} and refer to the hidden state at the top Transformer layer (before the output softmax) as \emph{hidden state} for brevity in this section\footnote{We also similarly analyzed unconditional NLG on the Yelp dataset~\citep{zhangCharacterlevelConvolutionalNetworks2015} and got the same conclusion. We leave the details in Appendix \ref{appb} due to the space limit.}.
\subsection{RC1: Formation of the Trap}
\label{subsec_rc1}
We decompose RC1 into three sub-claims and successively verify each of them.

\emph{Sub-claim 1: The output-layer embeddings of frequent tokens are more likely to approach corresponding hidden states during training}. Define the output embedding vectors as $\{\boldsymbol{w}_1, \boldsymbol{w}_2, \cdots, \boldsymbol{w}_V\}$ with vocabulary size $V$ and $\boldsymbol{w}_k \in \mathbb{R}^{1 \times d}$ corresponding to the $k$-th token $x_k$ where $d$ is hidden size, $c$ as the input condition (prompt). Following~\citep{RepDeg}, we consider the interaction of each word embedding and hidden state. During typical training, we aim to maximize the likelihood of the target token $x_i$  at the $t$-th time step:
\begin{align}
P(y_t\!=\!x_i|c, y_{<t}) & = \frac{\exp(\boldsymbol{w}_i\boldsymbol{h}^T_t)}{ \exp(\boldsymbol{w}_i\boldsymbol{h}^T_t)+C} \notag \\
& = 1-\frac{C}{\exp(\boldsymbol{w}_i\boldsymbol{h}^T_t)+C},
\label{eq_mle}
\end{align}
where $\boldsymbol{h}_t$ is the hidden state at $t$-th time step, $y_{<t}$ means the preceding $t\!-\!1$ tokens, and $C=\sum_{j, j \ne i} \exp(\boldsymbol{w}_{j}\boldsymbol{h}^T_t)$. By ignoring the terms irrelevant to $\boldsymbol{w}_{i}$, \textit{i.e.}, $C$, we can attribute the loss mainly to $\boldsymbol{w}_i\boldsymbol{h}^T_t$ and further factorize it as: 
\begin{equation}
    \boldsymbol{w}_i\boldsymbol{h}^T_i=\Vert \boldsymbol{w}_i\Vert\Vert\boldsymbol{h}_i \Vert \cos \theta.
\label{eq_cos}
\end{equation}

\begin{figure}[htp]
\center
\includegraphics[width=0.4 \textwidth]{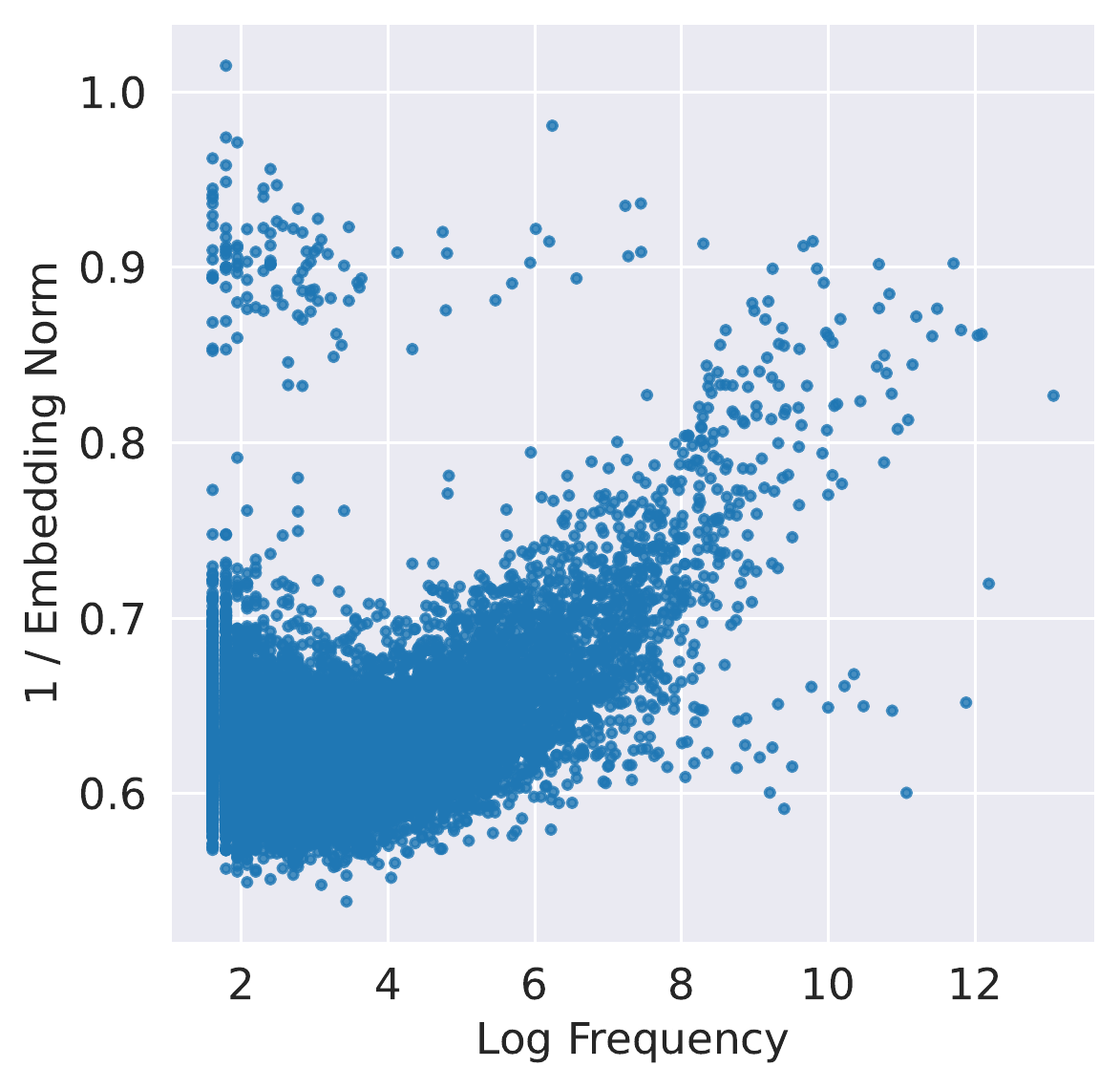}
\caption{The correlation between reciprocal embedding norm and logarithmic word frequency. The salient positive correlation ($R^2=0.835$) indicates that high-frequency words tend to have smaller embedding norms.}  
\label{fig_freq_roc} 
\end{figure}

From Eq.~(\ref{eq_cos}), we could derive that maximizing Eq.~(\ref{eq_mle}) actually minimizes the angle $\theta$ between $\boldsymbol{w}_i$ and $\boldsymbol{h}_t$. Since $\Vert\boldsymbol{h}_i \Vert$ is bounded by the layer normalization~\cite{pmlr-v119-xiong20b}, smaller $\Vert\boldsymbol{w}_i \Vert$ would require smaller $\theta$ to reach the same low loss. Based on this conclusion, we further observed that embeddings of high-frequency tokens tend to possess smaller norms, as shown in Fig.~\ref{fig_freq_roc}. As a result, we successfully verified sub-claim 1.

\begin{figure}[htp]
\flushright
\includegraphics[width=0.4\textwidth]{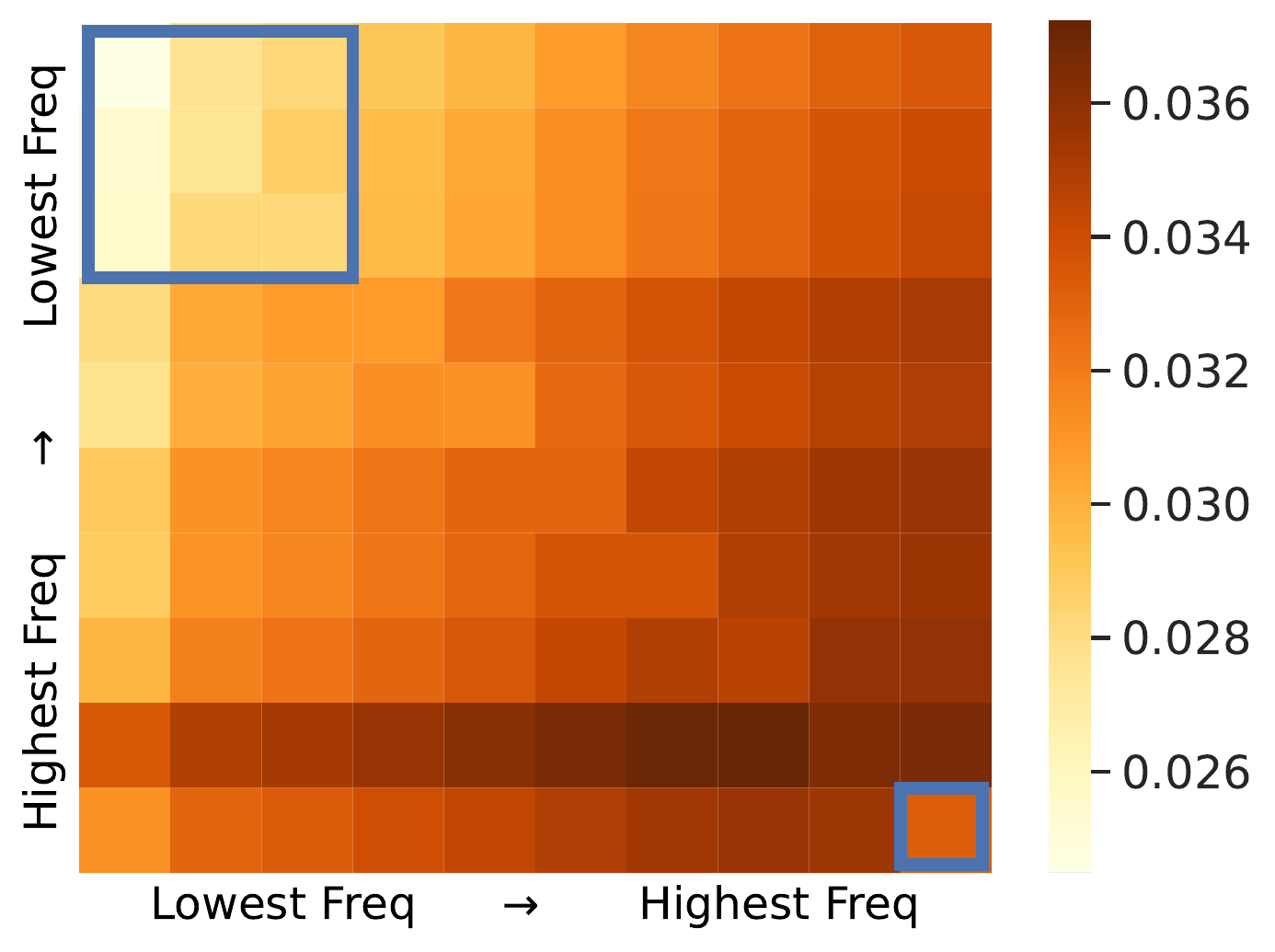}
\caption{The distance of embeddings in/between different frequency intervals. We divided token embeddings uniformly into 10 intervals by frequency and calculated the average $l_2$ distance between each embedding and the top-50 nearest ones in each interval.} 
\label{fig_mat_roc} 
\end{figure}

\emph{Sub-claim 2: The embeddings of high-frequency tokens tend to cluster}. We investigated the aggregation degree of token embedding, as shown in Fig.~\ref{fig_mat_roc}. It is obvious that rare tokens are clustered, as reported in~\cite{yu-etal-2022-rare}. Furthermore, we can find that the highly frequent (top 10\%) tokens are also closer to each other than the mid-frequency ones, which is also observed in BERT~\cite{li-etal-2020-sentence}, empirically supporting sub-claim 2. 

\emph{Sub-claim 3: There is a uniformly positive direction of the hidden states corresponding to the frequent tokens, \textit{i.e.}, $\boldsymbol{w_i}\boldsymbol{h}^T_{t}> 0$}. By slightly modifying the conclusion in~\citep{RepDeg}, we could demonstrate that there exists such a direction. See Appendix \ref{appc} for detailed derivation.

By combining these sub-claims, we could infer that embeddings of high-frequency tend to cluster and approach a uniformly positive direction of the hidden states during training, manifesting RC1.
\subsection{RC2: Falling into the Trap}
\label{subsec_rc2}
We have demonstrated how the trap is formed, which underlines the mediocre generated text. In this subsection, we further explore how the model falls into such a trap during the generation process.

In the Transformer self-attention, the hidden states $H^l$ in $l$-th layer are calculated as:
\begin{align}
\boldsymbol{h}^l_t =\text{softmax}\left(\frac{\boldsymbol{h}^{l\!-\!1}_t \boldsymbol{W}^{l}_q(\boldsymbol{H}^{l\!-\!1}\boldsymbol{W}^{l}_k)^T}{\sqrt{d}}\right)\boldsymbol{H}^{l\!-\!1}\boldsymbol{W}^{l}_v,
\label{eq_attn}
\end{align}
where $\boldsymbol{H}^{l-1}\!=\![\boldsymbol{h}^{l-1}_1;\cdots;\boldsymbol{h}^{l-1}_T]$ are hidden states of the previous layer and $\boldsymbol{W}^{l}_q,\boldsymbol{W}^{l}_k,\boldsymbol{W}^{l}_v$ are query, key and value projection matrix, respectively.

From Eq.~(\ref{eq_attn}), we can clearly see that each $\boldsymbol{h}_t$ would mix multiple previous hidden states by the attention score. We name such mixture as the \emph{attentive mixture}. Though in RC1 we only focus on hidden states corresponding to frequent target tokens, these high-frequency tokens are more likely to appear in the context of previous tokens, which could push the mixed $\boldsymbol{h}_t$ towards the uniformly positive direction and hence draw it closer to the cluster of high-frequency embeddings. As a result, the model falls into \emph{the Trap of Medicority}.
\begin{figure}[tp]
\centering
\includegraphics[width=0.4 \textwidth]{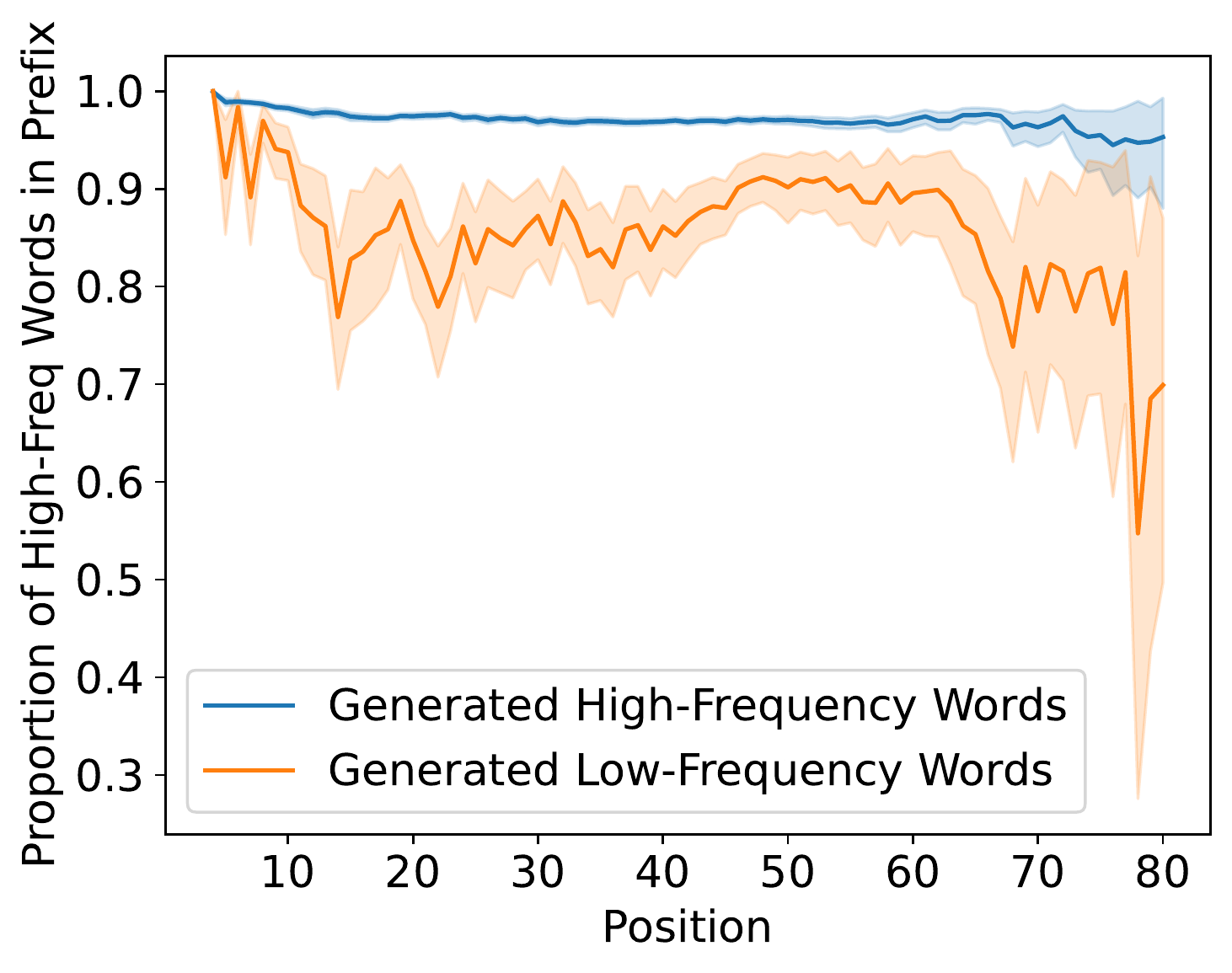}
\caption{The proportion of high-frequency words in the current sub-sequence when a high/low-frequency token is generated. The gap between the two lines indicates that low-frequency words tend to be generated from the context containing less high-frequency words. The sequences were generated by beam search.}
\label{fig_propo} 
\end{figure}
\begin{figure}[tp]
\centering
\includegraphics[width=0.4\textwidth]{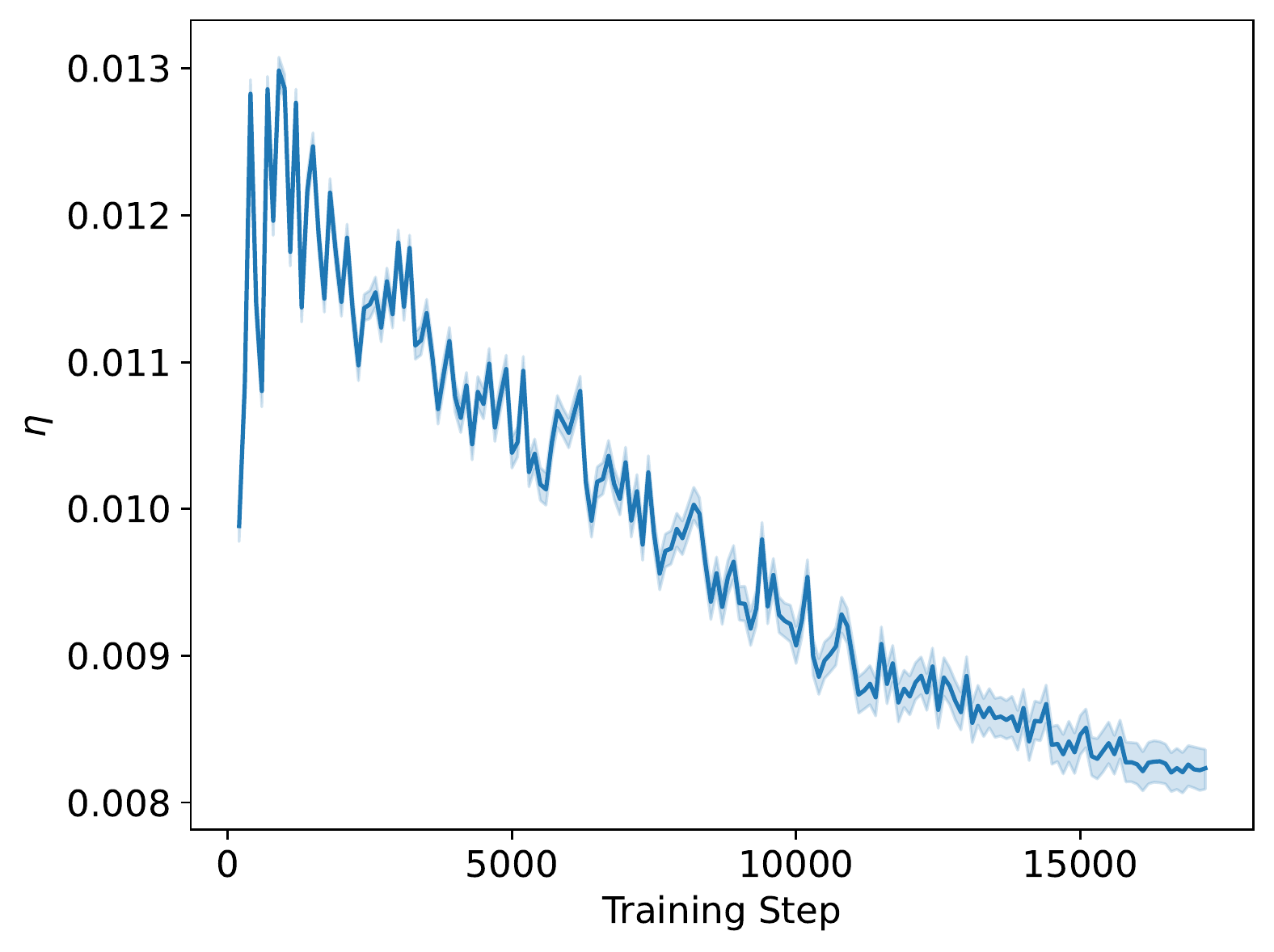}
\caption{The value of $\eta$ in Sub-claim 4 during training.} 
\label{fig_sanity_roc} 
\end{figure}

Moreover, as the sequence becomes longer, the number of high-frequency tokens also increases, which further aggravates the tendency of frequent tokens and keeps trapping the model. Fig.~\ref{fig_propo} manifests such deterioration. We can observe that with a longer sentence, GPT-2 requires increasing more infrequent tokens in context to motivate a low-frequency token. Inversely, the context with a large proportion of high-frequency tokens would always encourage frequent tokens at the next time step. Besides, we also calculated the pairwise cosine similarity between embeddings and the hidden states of all sentences in the testing set and then found that compared to the low-frequency embeddings, the high-frequency ones are generally nearer to all hidden states by a stable and large margin.

By these observations, we can empirically conclude that the attentive mixture process of hidden states forces a closer distance between each hidden state and frequent token embeddings, and hence causes higher generation probabilities (see Eq.~(\ref{eq_mle})) of high-frequency words, validating RC2.
\subsection{RC3: Dispersion dominates the mixture}
\label{subsec_rc3}
Subsections \ref{subsec_rc1} \& \ref{subsec_rc2} describe the complete process of how the trap of mediocrity forms and how the model falls into it. This process is dominated by the attentive mixture and would further deteriorate with a more scattered attention distribution.

To theoretically verify this claim, following~\cite{yu-etal-2022-rare}, we investigate the negative log-likelihood loss of a target sequence with length $T$:
\begin{equation}
\mathcal{L}_{NLL}=-\frac1T\sum_{t=1}^T \log\frac{\exp(\boldsymbol{w}_{x_t}\boldsymbol{h}^T_t)}{\sum_j^V \exp(\boldsymbol{w}_j\boldsymbol{h}^T_t)},
\end{equation}
where $x_t$ is the ground-truth token at the $t$-th step.

We rewrite the attention score used in Eq.~(\ref{eq_attn}) as:
\begin{equation}
 \boldsymbol{a}_{t} = \text{softmax}\left(\frac{\boldsymbol{h}_t \boldsymbol{W}^{l}_q(\boldsymbol{H}\boldsymbol{W}^{l}_k)^T}{\sqrt{d}}\right),
\end{equation}
where $\boldsymbol{a}_{t}=[a_{t,1},\cdots,a_{t,t}]$ represent the attention weights towards the preceding tokens.

For simplification, we ignore the nonlinearity in attention layers and let $\boldsymbol{h}^{l}_t=\sum_j a_{t,j}\hat{\boldsymbol{h}}_j$, where $\hat{\boldsymbol{h}}_j\!=\!\boldsymbol{h}^{l-1}_j \boldsymbol{W}^{l}_v$, and give the following conclusion:

\emph{Sub-claim 4:With the extremely dispersed attention distribution which reaches the highest entropy, \textit{i.e.}, for each time step $t$, $a_{t,j}=\frac{1}{t}, j \in \{1,\cdots,t\}$, the optimization direction of the embeddings would approach the uniformly positive direction of all hidden states if $\eta=\frac{1}{t}\sum_{j=1}^t\frac{p_j}{j}>0$ for each $t$, where $p_j$ is the generation probability of the ground-truth token $x_i$ of the $t$-th step, $p_t=p(y_t=x_i|c,y_{<t})$.}
\label{thm_sec2}

A theoretical demonstration of this sub-claim could be referred to Appendix \ref{appc}.

Sub-claim 4 indicates that when $\eta>0$, the improperly scattered attention would push embeddings towards the uniformly positive direction where high-frequency embeddings cluster, rather than more diverse directions, which is fundamental in forming the trap as described in Sec.~\ref{subsec_rc2}. We further examine $\eta$ during the training process and plot the result in Fig.~\ref{fig_sanity_roc}. Generally, $\eta$ is positive during the whole training process, proving RC3.

By combining RCs 1$ \sim $3, we successfully demonstrate that the improperly scattered attention is the underlying cause of poor diversity. A proper attention score should concentrate on some specific parts of the context, while in practice, the model would lose concentration and fall into the \emph{Trap of Medicority}, which implies we could break the trap and improve diversity by concentrating attention.
\section{Methodology}
\subsection{\textsc{Care}}
According to our analyses in Sec.~\ref{sec:analysis}, the improperly dispersed (low-sparsity) attention is one fundamental cause of the \emph{Trap of Medicority}. Therefore, we propose a novel regularization method, \textsc{Care}, to concentrate attention and thus promote diversity.

In detail, we add a regularization term to the original Maximum Likelihood Estimate (MLE) loss:
\begin{align}
    \mathcal{L}_R & = \sum_{t=1}^T\frac{\alpha(t+1)}{t(\alpha-1)} \Vert \hat{\boldsymbol{a}}_t\Vert_1,
\label{equ_lr}\\
    \mathcal{L}_{CARE} &= \mathcal{L}_{MLE} + \gamma \mathcal{L}_R,   
\end{align}
where $\hat{\boldsymbol{a}}_t$ denotes the unnormalized attention logits before softmax and attention dropout, $\Vert\cdot\Vert_1$ is the $l_1$ norm, $T$ is the sequence length, $\alpha > 1$ is a hyperparameter to control the sharpness of the attention distribution, and $\gamma$ could be utilized to balance the MLE term and the  regularization term.

To theoretically demonstrate our method's effectiveness in enhancing attention sparsity, we present the following conclusion:
\begin{theorem}
Minimizing $\mathcal{L}_R$ in Eq.~(\ref{equ_lr}) is equal to minimizing an upper bounds of the Rényi entropy\footnote{\url{https://en.wikipedia.org/wiki/R\%C3\%A9nyi\_entropy}} of the attention distribution $\mathrm{H}_{\alpha}(\boldsymbol{a}_t)$, where $\mathrm{H}_{\alpha}$ denotes the Rényi Entropy and $\alpha$ is the sharpness controller in Eq.~(6).
\label{thm2}
\end{theorem}

Theorem \ref{thm2} reveals that by minimizing $\mathcal{L}_R$, we could reduce the entropy of the attention distribution and hence make attention more concentrated, which helps evade the trap and boost diversity. Please refer to Appendix \ref{appc} for detailed proof.

Furthermore, $\mathcal{L}_R$ is self-adaptive benefiting from the derived weight $\beta=\frac{\alpha(t+1)}{t(\alpha-1)}$. As shown in Fig.~\ref{fig_propo}, longer generated sequences (larger $t$) exacerbate the trap, but larger $t$ also leads to a smaller $\beta$, which forces the model to reduce $\Vert\hat{\boldsymbol{a}}_t\Vert_1$ more to reach the same low loss and vice versa. This adaptability could mitigate the deterioration caused by length.

Besides, we could also consider the benefits of \textsc{CARE} from the perspective of \emph{Bayesian Inference}. Following~\cite{gal2016dropout}, we give another conclusion to validate such an advantage:

\begin{theorem}
When combined with the attention dropout, optimizing $\mathcal{L}_{CARE}$ can be regarded as minimizing $KL[q_\theta(\tilde{\boldsymbol{a}}_t)||p(\tilde{\boldsymbol{a}}_t|c, x_t, y_t)]$, that is, learning a Bayesian approximation $q_\theta(\tilde{\boldsymbol{a}}_t)$ of the true posterior attention $p(\tilde{\boldsymbol{a}}_t|c, x_t, y_t)$ at each time step, where $x_t, y_t$ denote the input and target tokens at the $t$-th step, respectively, and $\tilde{\boldsymbol{a}}_t$ denotes the unnormalized attention logits before softmax and after attention dropout,
\label{thm3}
\end{theorem}

\emph{Proof}. See Appendix C.

Theorem \ref{thm3} demonstrates another strong endorsement of our model: \modelname could act as a posterior estimator of the attention distribution, which helps improve the accuracy of attention and hence maintain comparable generation quality. In comparison, straightforward sparse attention methods without such theoretical guarantees can improve diversity but also hurt quality. See Table~\ref{res_conditional} for detailed results.

In addition, to conform to this theorem, we need to apply attention dropout to the unnormalized attention distribution, achieved by by sampling the dropping mask from a Bernoulli distribution and accordingly adding a large negative constant to the unnormalized attention logits. We also compare this setting with the
the original one in Transformer in Appendix \ref{appd} and manifested the superiority of ours.

Noth that \modelname is transparent to different model architectures since it involves only the output attention weights in Transformer, and hence can be easily implemented within 20 lines of codes. We provide the implementation details in Appendix~\ref{app_code}. 
\subsection{\textsc{Care}-A}
Since Theorem~\ref{thm3} indicates that the attention dropout is crucial for our method, we could further enhance the capacity of \modelname by incorporating the more flexible Concrete Dropout~\cite{DBLP:conf/nips/condrop}. The original dropout can be considered as sampling attention masks $z$ from a Bernoulli distribution $z\sim B(1,p)$ which could influence attention concentration through a specified dropout ratio $p$. In contrast, Concrete Dropout relaxes $z$ as $\tilde{z}\!=\!\sigma((\log p \!-\!\log (1\!-\!p)+\log u\!-\!\log (1\!-\!u)) / \tau )$, where $p$ is learnable, $\tau$ is the temperature, $\sigma$ means sigmoid, and $u \!\sim\! \text{Uniform}(0, 1)$. As the optimization requires an extra entropy term of $p$ as described in~\cite{DBLP:conf/nips/condrop}, we rewrite $\mathcal{L}_{CARE}$ as:
\begin{equation}
    \mathcal{L}_{CARE-A} = \mathcal{L}_{MLE} + \gamma \mathcal{L}_R + \delta \sum_{i=1}^L \mathrm{H}(p_i),
\end{equation}
where $L$ is the number of Transformer layers.

In this way, we learn layer-wise and adaptive dropout ratios $p$, benefiting a more flexible control of attention concentration. Therefore, we name this variant of our model as \textsc{Care}-A.

\section{Experiment}
\label{sec:experiment}

\begin{table*}[h]
\centering
\scalebox{0.8}{
\begin{tabular}{c|cccccccc|ccc}
\toprule
\multirow{2}{*}{Model} & \multicolumn{8}{c|}{Quality}                          &  \multicolumn{3}{c}{Diversity} \\ \cline{2-12}
                       & R-2$\uparrow$   & R-3$\uparrow$    & R-L$\uparrow$   & R-W$\uparrow$  & B-2$\uparrow$  & B-4$\uparrow$     & BS$\uparrow$     & CND$\downarrow$    &  Dist$\uparrow$      & JS$\downarrow$       & SB$\downarrow$      \\ \midrule
\multicolumn{11}{c}{Dataset: ParaSCI}\\
\midrule
GPT-2    & 41.48 & 32.65 & 54.97 & 35.28 & \textbf{41.64} & \underline{26.58} & 90.81 & 1.688 & 60.34 & 0.0726 & 17.52 \\
\hline
Pattern  & 41.46 & 32.58 & 55.06 & 35.34 & 41.31 & 26.43 & \underline{90.89} & 1.652 & 60.65 & 0.0683 & 16.86 \\
Entmax   & 38.44 & 29.78 & 52.28 & 33.50 & 38.06 & 23.60 & 90.35 & 1.769 & 59.30 & 0.0707 & 18.10 \\
$l_0$-Drop   & 37.26 & 28.64 & 50.99 & 32.53 & 37.96 & 23.47 & 90.05 & 1.700 & \underline{60.94} & \underline{0.0591} & \underline{16.18} \\
$\mathcal{L}_{\mathrm{A}}$-Tuning & 40.71 & 32.05 & 54.20 & 34.89 & 39.95 & 25.48 & 90.70 & \underline{1.632} & 60.67 & 0.0622 & 16.28 \\ \hline
\textsc{Care}    & \textbf{42.49} & \textbf{33.74} & \underline{55.69} & \textbf{35.94} & 41.59 & \textbf{26.75} & 90.80 & \textbf{1.631} & \textbf{61.04} & \textbf{0.0566} & \textbf{16.05} \\
\textsc{Care}-A & \underline{41.95} & \underline{33.01} & \textbf{55.79} & \underline{35.76} & \underline{41.63} & 26.56 & \textbf{91.04} & 1.637 & 60.60 & 0.0596 & 17.04\\
\midrule
\multicolumn{11}{c}{Dataset: ROCStories}\\
\midrule
GPT-2    & 6.700 & 1.321 & 23.17 & \underline{11.19} & 5.817 & 0.266 & 83.53 & 8.856 & 25.72 & 0.6625 & 62.36 \\
\hline
Pattern  & 6.690 & 1.301 & 23.33 & 11.09 & 6.207 & 0.273 & \underline{83.64} & 9.006 & 27.20 & 0.6280 & 61.16 \\
Entmax   & 6.011 & 1.011 & 22.42 & 10.80 & 5.374 & 0.150 & 82.91 & 9.521 & 25.03 & 0.7317 & 62.57 \\
$l_0$-Drop   & 5.850 & 1.065 & 22.11 & 10.77 & 4.896 & 0.177 & 83.17 & \textbf{8.489} & 24.82 & 0.6207 & 62.72 \\
$\mathcal{L}_{\mathrm{A}}$-Tuning & 6.367 & 1.214 & 22.94 & 10.96 & 5.809 & 0.226 & 83.37 & 8.904 & \underline{27.24} & \underline{0.5905} & \underline{61.06} \\ \hline
\textsc{Care}     & \underline{6.905} & \textbf{1.399} & \underline{23.54} & 11.08 & \textbf{6.829} & \textbf{0.326} & \textbf{83.74} & 9.735 & \textbf{27.77} & 0.6328 & \textbf{60.41} \\
\textsc{Care}-A   & \textbf{7.002} & \underline{1.332} & \textbf{23.80} & \textbf{11.32} & \underline{6.489} & \underline{0.274} & 83.63 & \underline{8.742} & 25.89 & \textbf{0.5708} & 61.51\\
\bottomrule
\end{tabular}
}
\caption{Evaluation results for conditional generation. The best results are \textbf{bolded} and the second best ones are \underline{underlined}. B, R, SB and BS represent BLEU, ROUGE, Self-BLEU and BertScore, respectively.}
\label{res_conditional}
\end{table*}
\subsection{Datasets}
We conducted comprehensive experiments on three conditional generation tasks: story generation on ROCStories~\citep{mostafazadeh-etal-2016-roc}, headline generation on MIND~\citep{wu-etal-2020-mind} and PENS~\citep{pens}, and paraphrase generation on ParaSCI~\citep{DBLP:conf/eacl/ParaSCI}. Specifically, we use the ArXiv set in ParaSCI which contains more instances. MIND and PENS are merged and re-split for more training data. For unconditional generation, we utilized the popular Yelp dataset~\citep{zhangCharacterlevelConvolutionalNetworks2015}. We truncated the text in these datasets to a max length of 1024 tokens. Detailed data statistics are listed in Appendix \ref{appa}.
\subsection{Implementation Details}
We use GPT-2 base~\citep{radford2019language} as the backbone for ROCStories, ParaSCI, and Yelp, and a little smaller model for headline generation due to the limit of our computational power. We train our \modelname, all baseline models, and the BPE tokenizers from scratch on each dataset. We use beam search for conditional generation datasets to avoid the diversity from sampling randomness, and sampling decoding for unconditional generation since beam search may bring too much duplicated content on Yelp. All models are implemented by HuggingFace library\footnote{https://huggingface.co/} and are trained from scratch. See Appendix \ref{appa} for more setting details. 
\subsection{Baselines}
Besides the original \textbf{(a) GPT-2}~\citep{radford2019language}, since our main claim is the correlation between attention concentration and generation diversity, we also compared several strong sparse attention based methods. \textbf{(b) Pattern}: the fixed attention pattern with local and sliding windows~\cite{sparseTrans}. \textbf{(c) Entmax}~\cite{correia-etal-2019-adaptively}: a learned adaptive sparse attention method. \textbf{(d) $l_0$-Drop}~\cite{L0Drop}: this method uses attention masks sampled from the hard concrete distribution to achieve sparsity and control sparsity degree by the $l_0$ norm loss on the mask. \textbf{(e) $\mathcal{L}_{\mathrm{A}}$-Tuning}~\cite{latuning}: a method which directly adds an entropy term of attention distributions to the training loss. All models share the same configuration for fair comparisons. More details of baseline models are provided in Appendix \ref{appa}.
\subsection{Metrics}

\emph{For the conditional generation tasks}, we evaluate the \textbf{quality} of the generated text by BLEU~\citep{papineni-etal-2002-bleu}, ROUGE~\citep{lin-hovy-2002-manual}, BertScore~\citep{bertscore}, and CND~\cite{CND}. We report BLEU-2,4 and ROUGE-2,3,L,W, which are most commonly used in the literature. For \textbf{diversity}, we assess Dist~\citep{li-etal-2016-diversity}, JS~\citep{JS}, and Self-BLEU~\citep{SelfBLEU}. We report the geometric mean of 1-gram to 4-gram for CND, Dist, JS, and Self-BLEU. \emph{For unconditional generation}, we report BLEU, MAUVE~\citep{pillutla-etal:mauve:neurips2021}, and CND for quality measurement. MAUVE is multiplied by 100 for convenience. For diversity, we take the same three metrics in the conditional generation.
\subsection{Results}
\subsubsection{Conditional Generation}
Table~\ref{res_conditional} presents the evaluation results on ROCStories and ParaSCI. We leave those of headline generation in Appendix \ref{appd} due to space limitations. We can see that compared to the naive GPT-2, most models with sparse attention improve generation diversity to some extent, except for Entmax which is relatively unstable due to its iterative estimation of the Tsallis entropy. Such results could well support our motivation and claim that the scattered attention causes the \emph{Trap of Medircority} and more concentrated attention helps evade the trap.

However, these sparse attention baseline methods significantly decrease the quality of generated text when optimizing diversity, while \textsc{Care}(-A) attained superior performance in novelty and diversity and meanwhile even boosted the quality of the generated text, demonstrating its effectiveness from Theorem \ref{thm3}. In addition, Pattern also increased the diversity without hurting quality, but is still inferior to our method under most metrics.

\subsubsection{Unconditional Generation}
\begin{table}[ht]
\centering
\resizebox{\columnwidth}{!}{
\begin{tabular}{c|cccc|ccc}
\toprule
\multirow{2}{*}{Model} & \multicolumn{4}{c|}{Quality}                          &  \multicolumn{3}{c}{Diversity} \\ \cline{2-8}
& B-2$\uparrow$  & B-4$\uparrow$    & CND$\downarrow$   & MV$\uparrow$ &  Dist$\uparrow$     & JS$\downarrow$       & SB$\downarrow$      \\ \midrule
GPT-2    &76.68 & 36.46 & \underline{0.886} & \underline{0.931} & 27.62 & 0.187 & 47.76 \\
Pattern  &76.96 & \underline{38.09} & 0.924 & 0.859 & 27.34 & 0.203 & 49.31 \\
Entmax   &77.03& 32.18 & 0.951 & 0.920 & 23.99 & 0.222 & 49.18 \\
$l_0$-Drop   &76.71 & 34.92 & \textbf{0.866} & 0.919 & 26.33 & 0.194 & 47.38 \\
$\mathcal{L}_{\mathrm{A}}$-Tuning &\underline{77.35} & \textbf{38.66} & 0.910 & 0.835 & 26.26 & 0.211 & 50.82 \\ \hline
\textsc{Care}     &73.84 & 33.61 & 1.033 & 0.863 & \textbf{30.84} & \underline{0.178} & \textbf{43.60} \\
\textsc{Care}-A & \textbf{77.53} & 35.69 & 0.933 & \textbf{0.974} & \underline{28.44} & \textbf{0.169} & \underline{44.65}\\
\bottomrule
\end{tabular}
}
\caption{Evaluation results for unconditional generation on the Yelp dataset. MV stands for MAUVE.}
\label{res_unconditional}
\end{table}
The results of unconditional generation are presented in Table~\ref{res_unconditional}. In this task, most sparse baselines perform no better than GPT-2 on diversity. This phenomenon could be attributed to the sufficient diversity brought by the sampling decoding. However, our method can still generate more diverse content compared to GPT-2, manifesting its extreme power in evading mediocrity and enhancing novelty. \textsc{Care}-A even achieves comparable quality while significantly improving diversity. 
\subsection{Human Evaluation}
\begin{figure}[ht]
\centering
\includegraphics[width=0.4\textwidth]{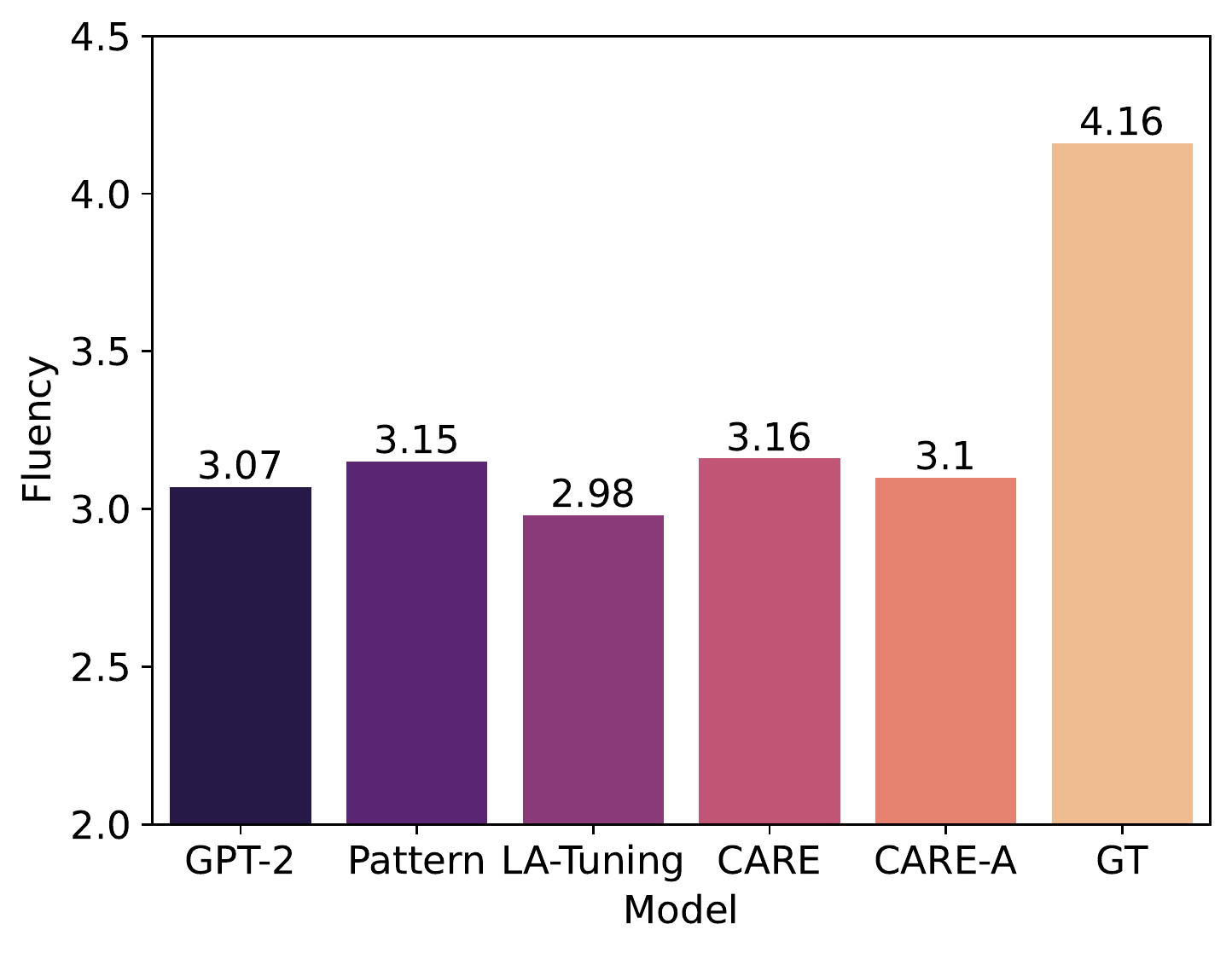}
\caption{Human evaluation on the fluency of the generated text. GT stands for human-written ground truth. The p-value $<$ 0.001, and the Kappa score is 0.67, indicating an acceptable inter-annotator agreement.} 
\label{human_eval} 
\end{figure}

\begin{figure*} [t]
	\centering
	\subfloat[\label{fig:a}]{
		\includegraphics[scale=0.33]{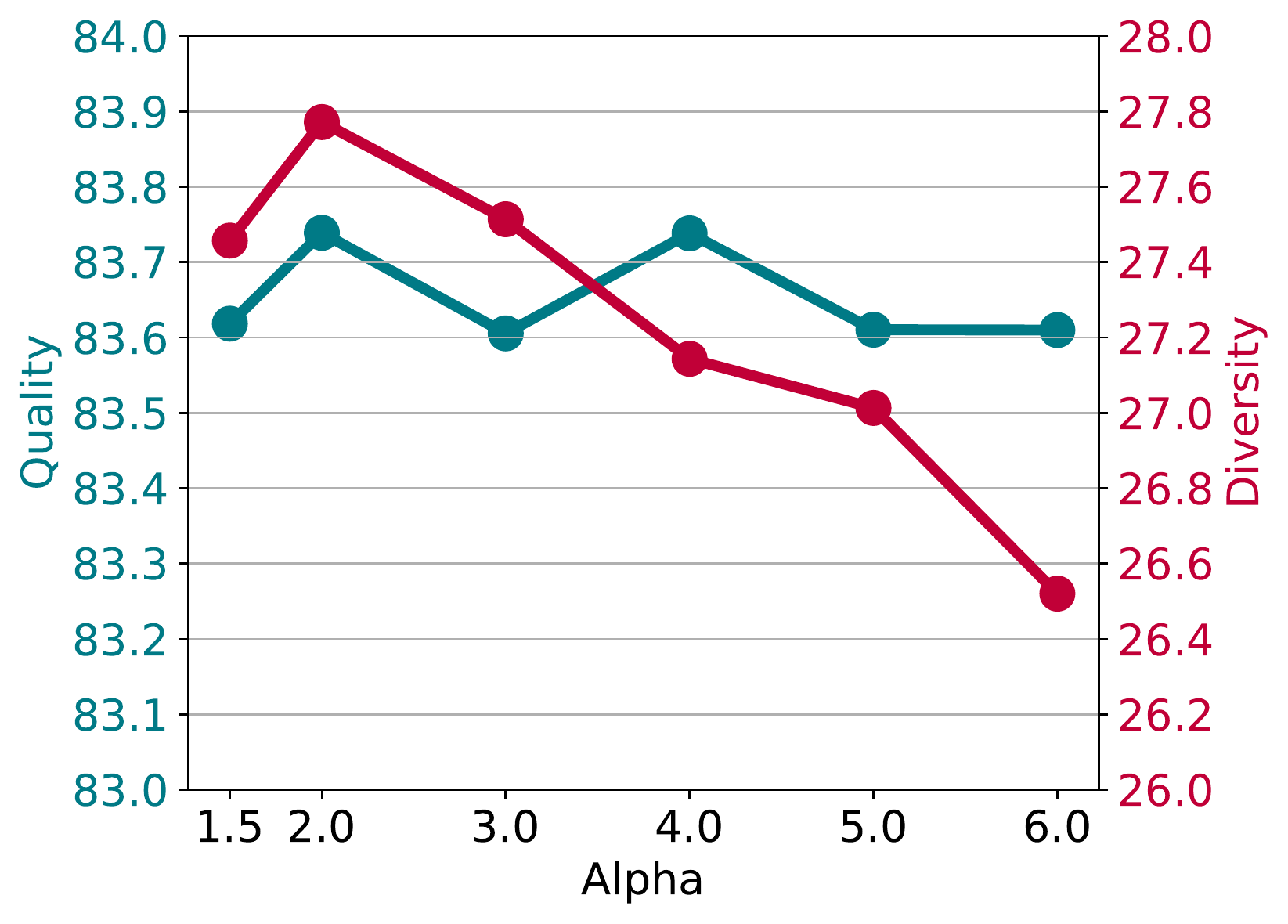}}
	\subfloat[\label{fig:b}]{
		\includegraphics[scale=0.33]{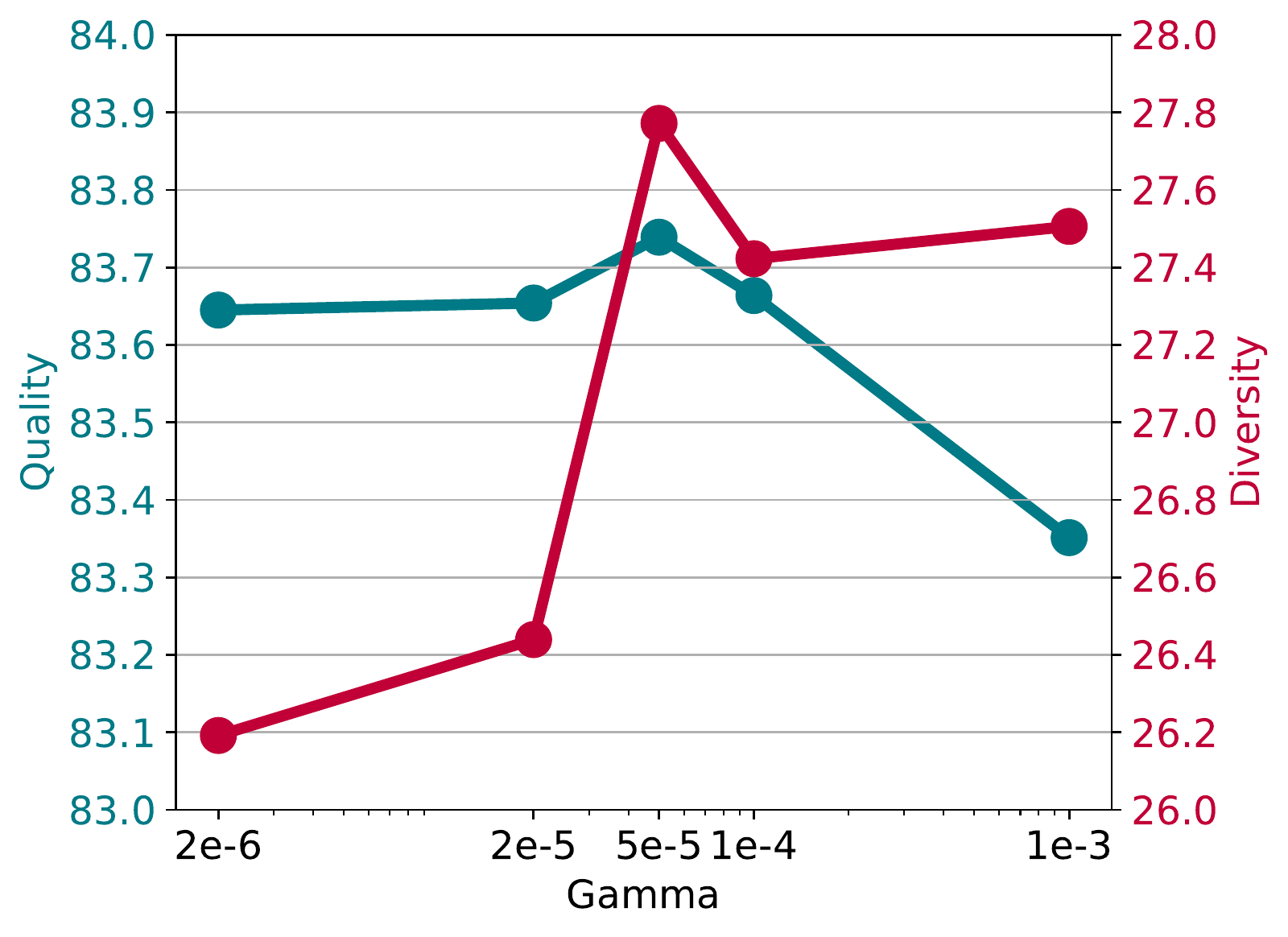}}
	\subfloat[\label{fig:c}]{
		\includegraphics[scale=0.33]{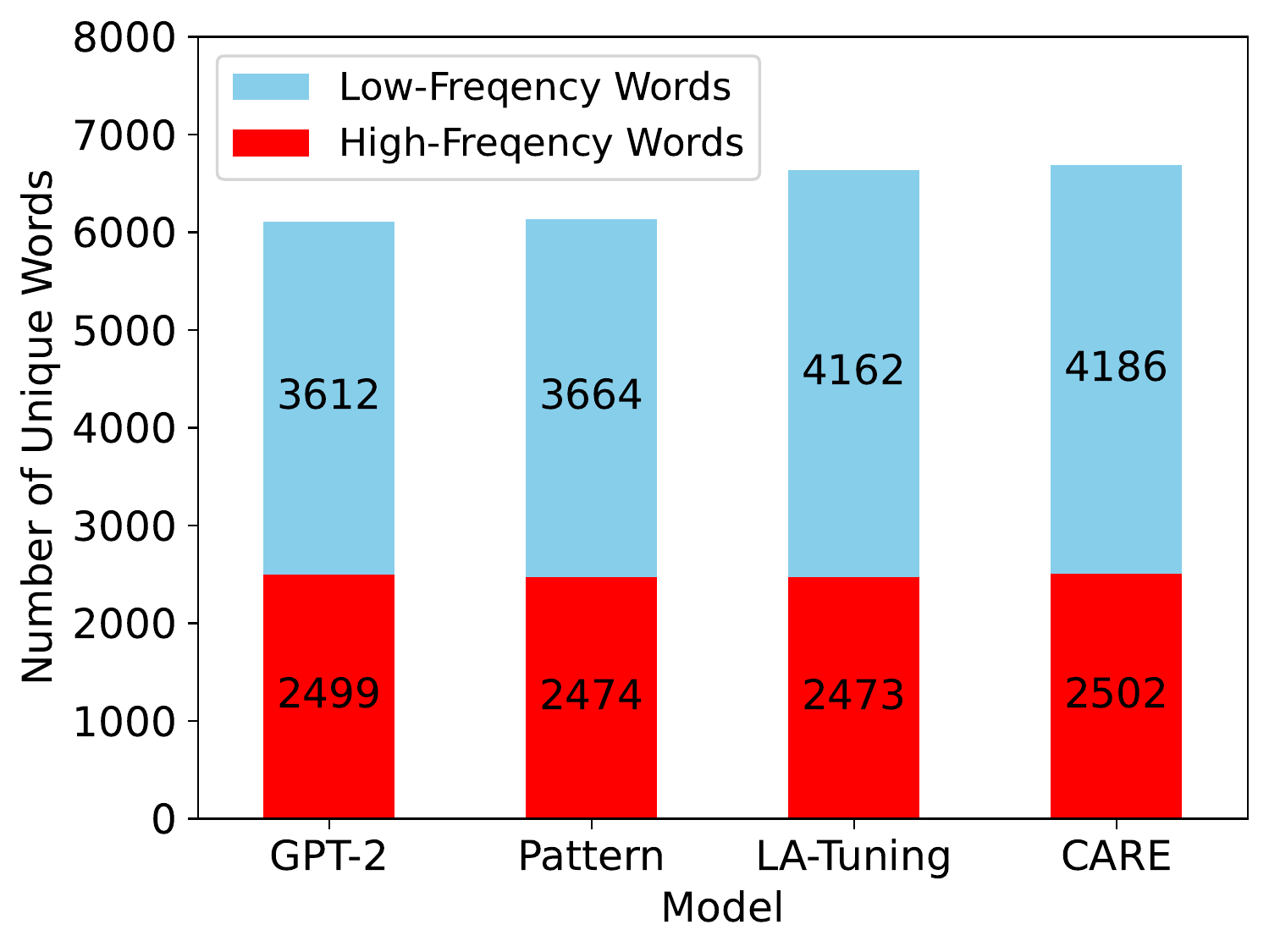}}
	\caption{(a, b) The performance when varying different $\alpha$ and $\gamma$. The quality of the generated text is measured by BERTScore and diversity by Dist. (c) The unique high-frequency words in the generated text of each model.}
	\label{fig3} 
\end{figure*}
We also conduct a human evaluation of the fluency of generated texts on ROCStories. See Appendix~\ref{appa:human} for the concrete evaluation protocol. As shown in Table~\ref{human_eval}, the fluency of the text generated by \textsc{Care} is comparable with GPT-2 and other baseline methods. Combined with the automatic evaluation, we could conclude that our model obtains better diversity without losing any fluency.

\subsection{Ablation Study}
To further investigate the effect of $\mathcal{L}_R$, we vary different values of $\alpha$ and $\gamma$ on ROCStories and present the results in Fig.~\ref{fig3}. Fig.~\ref{fig:a} shows that our model benefits from the adequate sharpness (larger $\alpha$) by the entropy regularization, while excessively sharp or blunt attention hurts both quality and diversity. Besides, the weight of $\mathcal{L}_R$ also needs to be carefully tuned since an overly small weight cannot exert the effect on enhancing sparsity. In contrast, an unduly large weight (too large $\gamma$) would hinder the optimization of MLE and result in a decline in generation quality, as shown in Fig.~\ref{fig:b}.

\subsection{Novelty Analysis}
To verify whether our method achieves better novelty by inclining to the low-frequency words, we separately count the unique high/low-frequency words generated by \textsc{Care} in ROCStories following ~\citep{DBLP:conf/acl/YuSK0RY22raredege}. Please note the subtle difference between diversity and novelty. Only sufficient novel (distinct) words can support satisfactorily good inter-instance diversity. As presented in Fig.~\ref{fig:c}, the result demonstrates that our model generates more unique low-frequency words than GPT-2 and Pattern by a large margin and $\mathcal{L}_{\mathrm{A}}$Tuning by a small margin, which indicates that the diversity and novelty brought by sparse attention could be attributed to the incline of low-frequency words. This finding again confirms that the diversity improvement comes from evading the \emph{Trap of Mediocrity}, while our method is the most successful Escapist.
\subsection{Case Study}
\begin{figure}[htbp]
    \centering
    \includegraphics[scale=0.35]{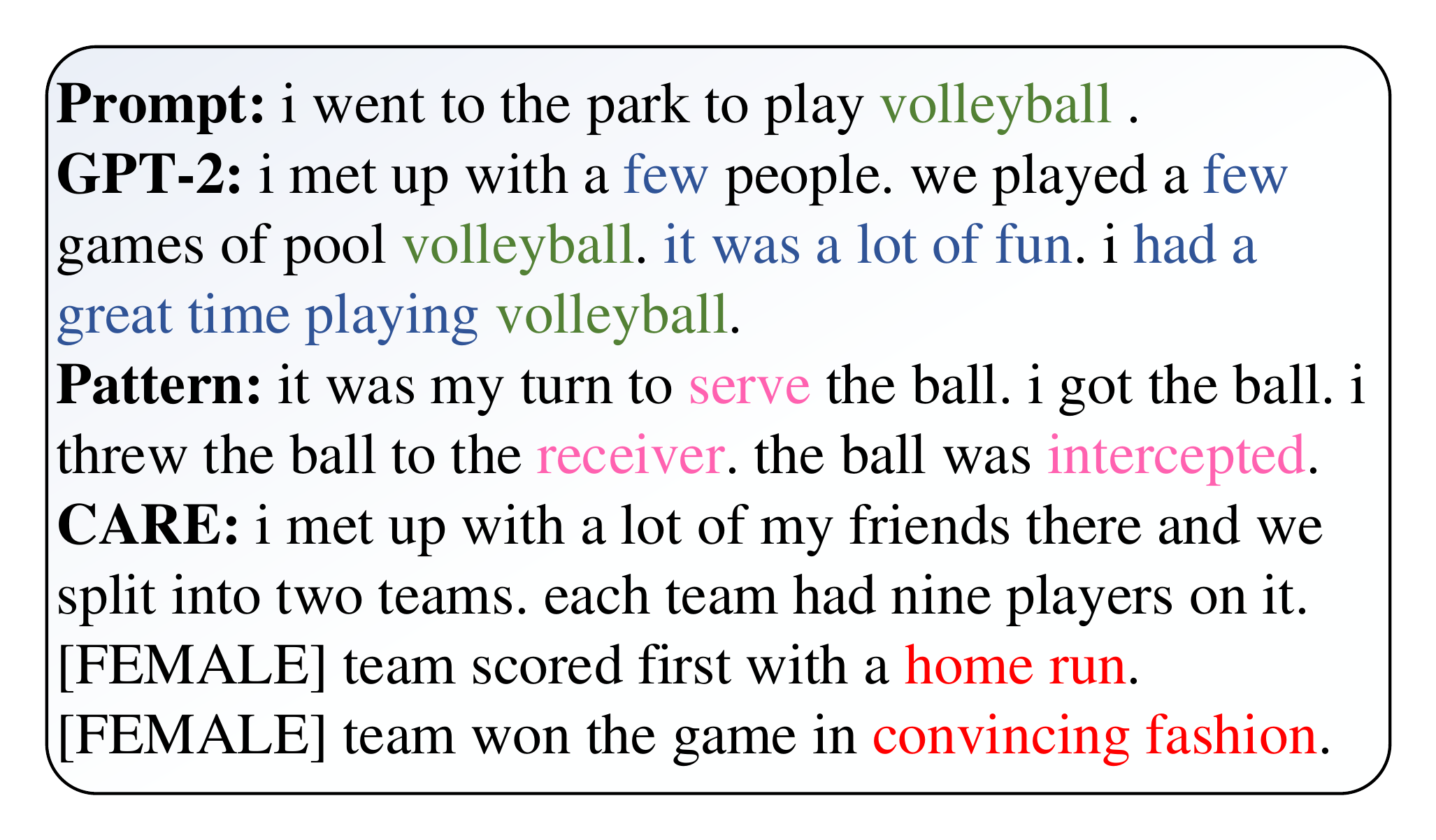}
    \caption{A generated sample in ROC. \textcolor{blueviolet}{Generic} / \textcolor{darkgreen}{repetitive} continuations are marked in different colors, while those in \textcolor{reallypink}{pink} / \textcolor{red}{red} are relevant/novel expressions.}
    \label{fig:case}
\end{figure}
Fig.~\ref{fig:case} shows a typical sample generated by \textsc{Care} and the baselines. GPT-2 generates relatively generic continuations with mediocre descriptions like \emph{it was a lot of fun} and simply copies the word \emph{volleyball} several times, while models with sparse attention could generate novel and rich content precise to the given topic. \textsc{Care} could further generate more novel and topic-related phrases like \emph{home run} and \emph{convincing fashion}.
\section{Related Work}
\paragraph{Enhancing Diversity in NLG} \quad Previous work focusing on diversity in text generation mainly falls into two lines. The first line eschews this issue by incorporating randomization and diversification into decoding algorithms, \textit{e.g.}, top-k~\citep{DBLP:conf/acl/LewisDF18topk} and top-p~\citep{curiousdege} sampling, diversified beam search~\citep{DBLP:journals/corr/LiMJ16DiverseBeam, DBLP:conf/aaai/VijayakumarCSSL18DiverseBeam}, and stochastic beam search~\citep{DBLP:conf/icml/KoolHW19}. The other line aims to substitute or supplement the MLE loss with novel objectives, which involves Reinforcement Learning~\citep{DBLP:conf/ijcai/ShiCQH18RLDiver}, VAE~\citep{DBLP:conf/iclr/NieNP19VAEDiver, hu-etal-2022-fuse}, adversarial training~\citep{DBLP:conf/acl/DiaoSSSZ21GANDiver, DBLP:conf/acl/ZhouLL20AIALDiver}, unlikelihood training~\citep{DBLP:conf/iclr/WelleckKRDCW20}, energy-based models~\citep{DBLP:conf/iclr/DengBOSR20Energy}, and additional penalty terms~\citep{DBLP:journals/corr/abs-2202-06417CTG, DBLP:journals/corr/abs-2206-02369BL}. There are also works that modify attention to handle this problem~\citep{Sun_Huang_Wei_Dai_Chen_2020multihead, DBLP:conf/acl/DongBLHBCC21Modu}. However, they only consider on-the-fly modification instead of analyzing the comprehensive relationship between attention and NLG diversity.

\paragraph{Sparse Attention} \quad There have been continuous efforts to reduce the computation cost of self-attention in Transformer. The majority make the dense attention matrix sparser to improve efficiency, including fixed patterns~\citep{sparseTrans, DBLP:journals/corr/abs-2004-05150longformer, DBLP:conf/emnlp/AinslieOACFPRSW20ETC, DBLP:conf/nips/ZaheerGDAAOPRWY20BigBird, DBLP:conf/acl/SukhbaatarGBJ19AdaSpan, HEPOS}, cluster-based methods~\citep{DBLP:journals/corr/abs-2009-06097Cluster, DBLP:journals/tacl/RoySVG21Routing}, learnable patterns~\citep{correia-etal-2019-adaptively, DBLP:conf/iclr/KitaevKL20Reformer, DBLP:conf/icml/TayBYMJ20Sinkhorn}, and regularization-based methods~\citep{L0Drop, DBLP:conf/icml/ShiGRXLLK21SparseBERT}. More details can be referred to~\citep{DBLP:journals/corr/abs-2009-06732survey}. However, all these methods focus on saving computational budget and improving the performance of language modeling. We are the first to study sparse attention from the in-depth view of improving generation diversity.
\section{Conclusion}
In this paper, we analyze the underlying cause of generic generated texts and propose a theory called the \emph{Trap of Mediocrity} to attribute this problem to the attention dispersion in Transformer. To evade such a trap, we present \textsc{Care} to concentrate attention, which is easy to implement and transparent to model structures. Experiments on a variety of NLG tasks demonstrate \textsc{Care}'s superiority in improving diversity while maintaining competitive quality. The code of our method will be released in \href{https://github.com/peterliwenhao/CARE}{https://github.com/peterliwenhao/CARE}.

\section*{Acknowledgement}
We sincerely thank all the anonymous reviewers for their valuable comments. This work is supported by the National Key R\&D Program of China (No. 2020AAA0106502) and Institute Guo Qiang at Tsinghua University.

\section*{Limitation}
There are still several limitations to our methods. Firstly, due to the introduction of new hypermeters, our method may need extra hyperparameter tuning when adapted to the specific task. Secondly, the validity of our approach to finetuning models pretrained with full attention has not been tested, which needs further experiments. Since we are the first to propose the \emph{Trap of Medicority} to our best knowledge, our exploration of this problem is inevitably very preliminary or even defective. We will keep refining our methods and expect more future work to further explore this problem. 

\bibliography{anthology,custom}
\bibliographystyle{acl_natbib}

\appendix

\onecolumn

\newcommand{\dif}{\mathop{}\!\mathrm{d}}

\section*{Appendix}

\section{Experimental Details}
\label{appa}

\subsection{Datasets}
We conduct our experiments on four datasets, ROCStories, ParaSCI, Yelp, and MP (combing MIND \& PENS). The statistics of the datasets are listed in Table~\ref{Training_Time}, and the GPU hours for training and inference are in Table~\ref{Statistics}. All the datasets are in English.

\textbf{ParaSCI} \citep{DBLP:conf/eacl/ParaSCI} A paraphrasing dataset collected from scientific literature, particularly ACL Anthology and arXiv. We use the ParaSCI-arXiv set since it contains more instances while following the split of their original paper.

\textbf{Yelp} \citep{zhangCharacterlevelConvolutionalNetworks2015} A dataset for unconditional generation collected from the food review on the Yelp website. We also follow the split released publicly.

\textbf{ROCStories} \citep{mostafazadeh-etal-2016-roc} A story corpus consists of 5-sentence short stories. We download it from the original website and randomly split it with the proportion in~\ref{Statistics}, following the previous papers \citep{JianGuan}. We use the first sentence as the condition to generate the following four sentences, which also agree with the settings in \citep{JianGuan}

\textbf{MP (MIND and PENS)} \citep{wu-etal-2020-mind,pens} They are two datasets consisting of the main body and the headline of online news, providing a source for generating the headline of news given its content. Since they are both organized from MSN News, we merged them for a larger dataset and divided the training, validation, and test dataset with the proportions in~\ref{Statistics}. We truncate the news body to 1024 tokens since some are too long.

\begin{table}[h]
\centering
\begin{tabular}{c|ccc}
\toprule
           & Train & Inference &  \\\midrule
ROCStories & 8h    & 7h        &  \\
ParaSCI    & 5h    & 3h        &  \\
MP         & 50h   & 5h        &  \\
Yelp       & 15h   & 2h        & \\
\bottomrule
\end{tabular}
\caption{The training and inferring time consumed for each dataset. The inference time means the time consuming of the inference process of one checkpoint on the whole test dataset for one round with the corresponding decoding strategy (sampling for Yelp and beam search for others).}
\label{Training_Time}

\end{table}
\begin{table}[h]
\centering
\begin{tabular}{c|ccc}
\toprule
Dataset & \#Train & \#Dev & \#Test \\\midrule
ROCStories & 88345   & 4908  & 4908   \\
ParaSCI    & 309834  & 3680  & 2549   \\
MP         & 214964  & 5000  & 10000  \\
Yelp       & 100000  & 10000 & 10000\\
\bottomrule
\end{tabular}
\caption{Dataset Statistics (Train/Dev/Test Split) for all the dataset we train on}
\label{Statistics}
\end{table}
\subsection{Metrics}

We use the following metrics to evaluate the quality of the generated text:

\textbf{BLEU} \citep{papineni-etal-2002-bleu} is a widely used metric in a variety of tasks on machine translation and language generation, calculating the n-gram overlap between the generated text and its corresponding reference. We use the implementation in the Huggingface Dataset Library\citep{DBLP:conf/emnlp/LhoestMJTPPCDPT21}. We reported BLEU-2 and BLEU-4 since they are most commonly used in language generation. In the unconditional generation task, we treat all the instances in the test set as references.

\textbf{ROUGE} \citep{lin-hovy-2002-manual} is another metric based on the n-gram overlap between the generated and referenced sequences, mostly used in automatic summarization. We also use the version from Huggingface Dataset Library\citep{DBLP:conf/emnlp/LhoestMJTPPCDPT21}. We measured ROUGE-2,3, L, and W on the generated texts.

\textbf{BertScore}\citep{bertscore} is a model-based metric measuring the semantic similarity by the cosine similarity between generated sequence and reference using pre-trained BERT contextual representations. We use the \emph{bert\_score} library\footnote{https://pypi.org/project/bert-score/} in our evaluation.

\textbf{CND} \citep{CND} is a distribution-level metric that approximates the divergence between the distribution of generated text and its corresponding reference in n-gram spaces. We use our self-implemented version of it.

\textbf{MAUVE} \citep{pillutla-etal:mauve:neurips2021} is another distributional-level metric measuring the distributional divergence between the generated text and the reference text. We use its official implementation from GitHub\footnote{\url{https://www.google.com/url?sa=t&rct=j&q=&esrc=s&source=web&cd=&cad=rja&uact=8&ved=2ahUKEwixy47LvMj4AhUeJ0QIHc-4CiUQFnoECAgQAQ&url=https\%3A\%2F\%2Fgithub.com\%2Fkrishnap25\%2Fmauve&usg=AOvVaw2PEycTPemGnrAHnGagzJEF}}.
 
For diversity and novelty, we take the following three metrics:

\textbf{JS} \citep{JS} measures the diversity of generated text by calculating the Jaccard Similarity between the n-grams of each pair of generated sequences. We implement it by ourselves.

\textbf{Dist} \cite{li-etal-2016-diversity} is a widely used metric for the novelty of the generated text, which calculates the proportion of distinct n-grams within the generated sentences. We also use the self-implemented version of it.

\textbf{Self-BLEU} \citep{SelfBLEU} measures the diversity of generated corpus by the average of pairwise BLEU score in the corpus. We use the version from the fast\_bleu library\footnote{\url{https://www.google.com/url?sa=t&rct=j&q=&esrc=s&source=web&cd=&cad=rja&uact=8&ved=2ahUKEwjc1NjWvcj4AhVTKEQIHXh-ADwQFnoECAkQAQ&url=https\%3A\%2F\%2Fgithub.com\%2FDanial-Alh\%2Ffast-bleu&usg=AOvVaw0Dun5WnEw2LscEz7Gyhxeu}}.

We reported the geometric mean of 1-4 grams for CND, JS, Dist, and Self-BLEU.

\subsection{Implementation Details}
We use GPT-2 base~\citep{radford2019language} as the backbone for ROCStories, ParaSCI, and Yelp, and a little smaller model for headline generation due to the limit of our computational power. The specific hyperparameter of the architecture of the model we demonstrate in ~\ref{model_param}. Models on the MP dataset is trained on NVIDIA GeForce GTX TITAN X, while models on the other three datasets is trained on NVIDIA GeForce RTX 3090.

For the datasets of conditional generation, we only compute loss on the continuation text but not on the conditional prefix. In addition, we train a BPE tokenizer for each dataset with the vocabulary size of 30000 to perform better tokenization.

We use beam search on the conditional generation dataset due to its stability and sampling for unconditional generation since beam search in this task tends to generate duplicated candidates. For the specific parameters in the unconditional generation, we use a hybrid strategy with top-p=0.9~\citep{curiousdege} and top-k=50~\citep{DBLP:conf/acl/LewisDF18topk} with temperature 1. We also perform a linear warmup on the $\mathcal{L}_R$ term in \textsc{Care} and the entropy term in \textsc{Care}-A. Our methods and all the baselines are implemented by HuggingFace Transformers library \citep{wolf-etal-2020-transformers}. All the models are trained from scratch.

\begin{table}[h]
\centering
\begin{tabular}{c|cccl}
\toprule
       & hidden\_size & head\_num & layer\_num & Total\_Parameters \\ \midrule
MP     & 512          & 8         & 8          & 39.33M            \\
Others & 768          & 12        & 12         & 101.03M   \\  \bottomrule
\end{tabular}
\caption{The detailed architecture of the model}
\label{model_param}
\end{table}
\subsection{Hyperparameter Tuning}

We ran the model for different epochs reported in Table~\ref{hyperparam_run}. We evaluate each epoch and report the best epoch due to the performance on the validation set.
We search each model on each dataset for ~30 hyperparameter trials. With two variants and four datasets, we ran our model about 240 times. We manually tune the models based on the overall performance of the model on the quality and diversity measurements of the validation set, mainly considerating CND, BertScore, Dist, and Self-BLEU. The upper and lower bounds of the hyperparameters of our model are demonstrated in Table~\ref{hyperparam}.
\begin{table}[h]
\centering
\begin{tabular}{c|c}
\hline
Dataset & Epoch \\ \hline
MP      & 20    \\
ROC     & 25-28 \\ 
ParaSCI & 5     \\ 
Yelp    & 25-30 \\ \hline
\end{tabular}
\caption{Running Epoch for different datasets}
\label{hyperparam_run}
\end{table}
\begin{table}[h]
\centering
\begin{tabular}{c|cc}
\hline
Hyperparameters                    & lower bound & upper bound \\ \hline
$\alpha$                           & 1.5         & 6.0         \\
$\gamma$                           & 1e-6        & 1e-2        \\
$\delta$                           & 1e-5        & 1e-3        \\
Learning Rate                      & 5e-5        & 5e-3        \\
Freezing Steps for $\mathcal{L}_R$ & 0           & 20000       \\
Warmup Steps for $\mathcal{L}_R$   & 0           & 20000       \\ \hline
\end{tabular}
\caption{Bounds for the hyperparameter used}
\label{hyperparam}
\end{table}

\subsection{Human Evaluation}
\label{appa:human}
We randomly selected 100 prompts from the ROCStories dataset and let each model generate corresponding continuations. In the annotation sheet, we present the prompt and the continuations to the annotators in the format of 100 groups. Each group contains six continuations generated from the same prompt by the five models and the ground truth. The order of the model (or ground truth) within each group is randomly shuffled, and the model each continuation from is unknown to the annotator. The different orders in different groups could alleviate the bias of the annotators. We recruited four annotators with adequate English skills and asked them to score the fluency of the shuffled stories on a scale of 1-5 and calculate the average scoring for each model, as shown in Figure \ref{human_eval}.

\section{Supplemental Materials for the Analysis of the \emph{Trap of Medicority}}
\label{appb}

\subsection{The Evidence that the Hidden State is nearer with the High-Frequency Word Embeddings}

\begin{figure}[tp]
\centering
\includegraphics[width=0.5\textwidth]{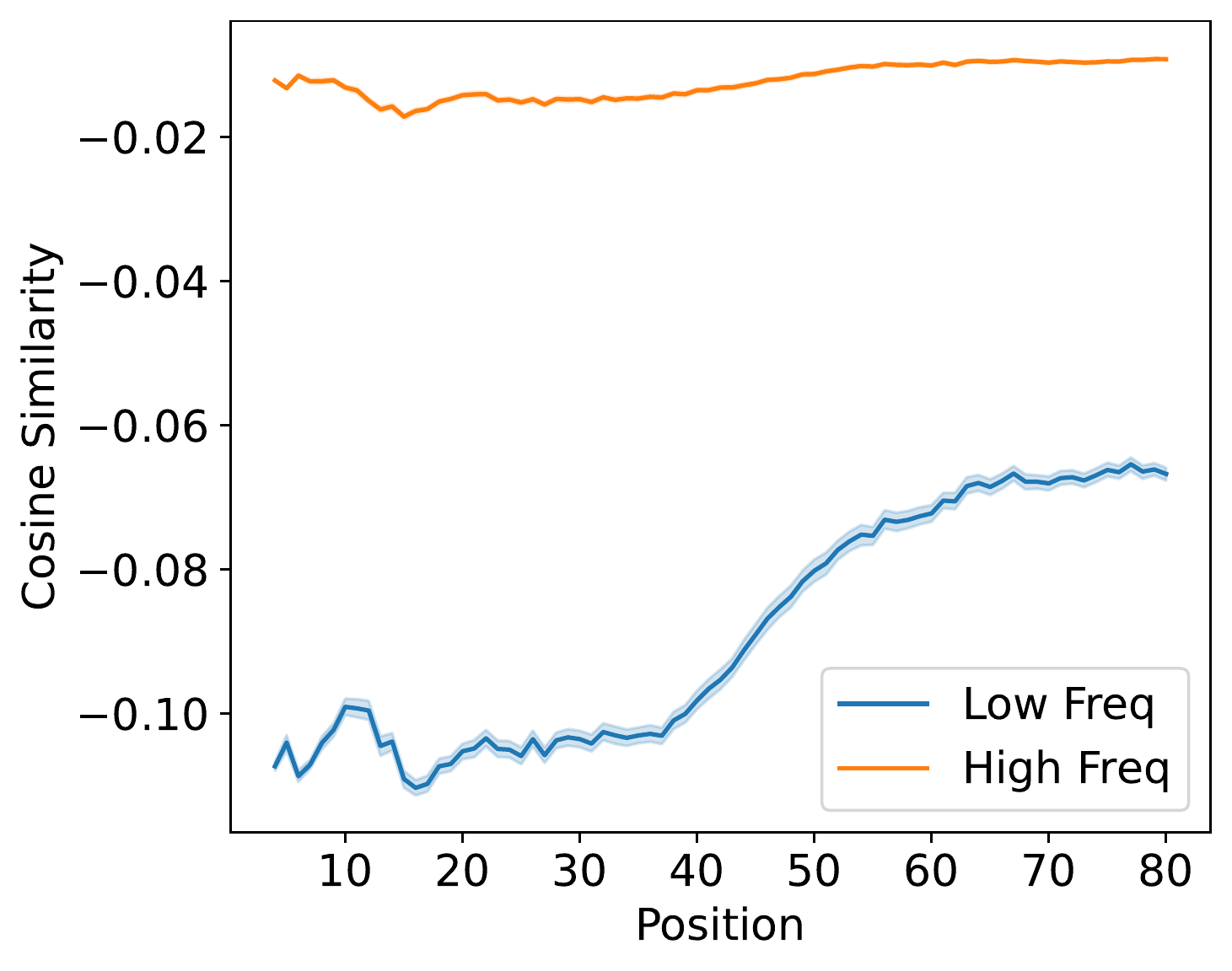}
\caption{The pairwise cosine similarity between the hidden state when generation and the high-frequency and low-frequency word embeddings. The gap between two lines indicated that the high-frequency word embeddings are closer to the hidden states} 
\label{cosdic_roc} 
\end{figure}

To prove the claim that the hidden states are closer to the high-frequency words, we calculate the pairwise cosine similarity between the hidden state when generation and the high-frequency and low-frequency word embeddings and demonstrate the result in Figure~\ref{cosdic_roc}. The significant gap between these two lines proves our assumption that the high-frequency word embeddings are closer to the hidden states than the low-frequency ones.

\subsection{The Addition Results on the Yelp Dataset}

\begin{figure}[ht]
\center
\includegraphics[width=0.5 \textwidth]{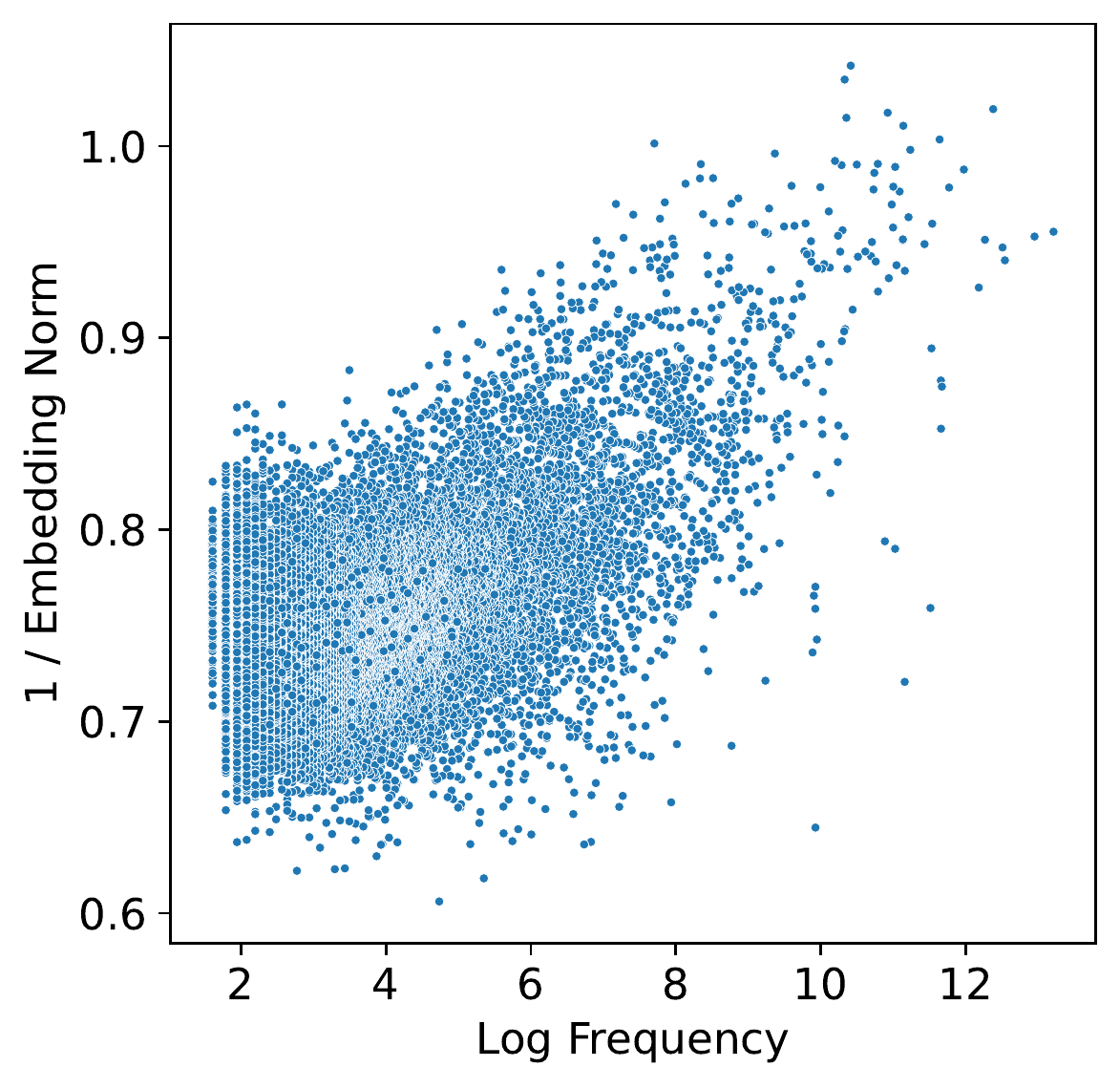}
\caption{The correlation between reciprocal embedding norm and logarithmic word frequency on the Yelp Dataset, with the same setting in Figure~\ref{fig_freq_roc}. The corresponding $R^2$ is 0.854.} 
\label{fig_freq_yelp} 
\end{figure}

\begin{figure}[ht]
\center
\includegraphics[width=0.5 \textwidth]{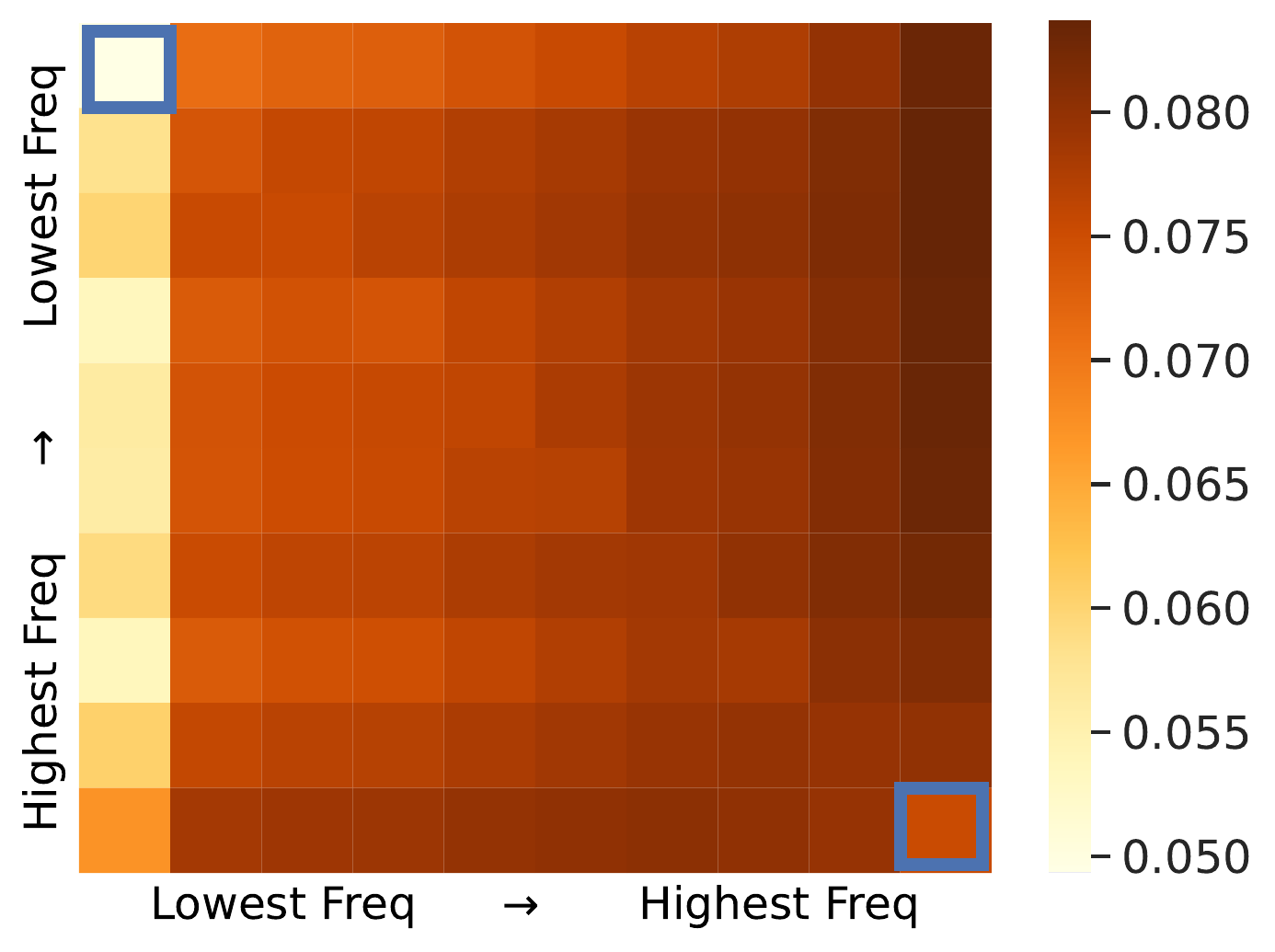}
\caption{The distance of embeddings in/between different frequency intervals on the Yelp Dataset, with the same setting in Figure~\ref{fig_mat_roc}} 
\label{fig_mat_yelp} 
\end{figure}

\begin{figure}[ht]
\center
\includegraphics[width=0.5 \textwidth]{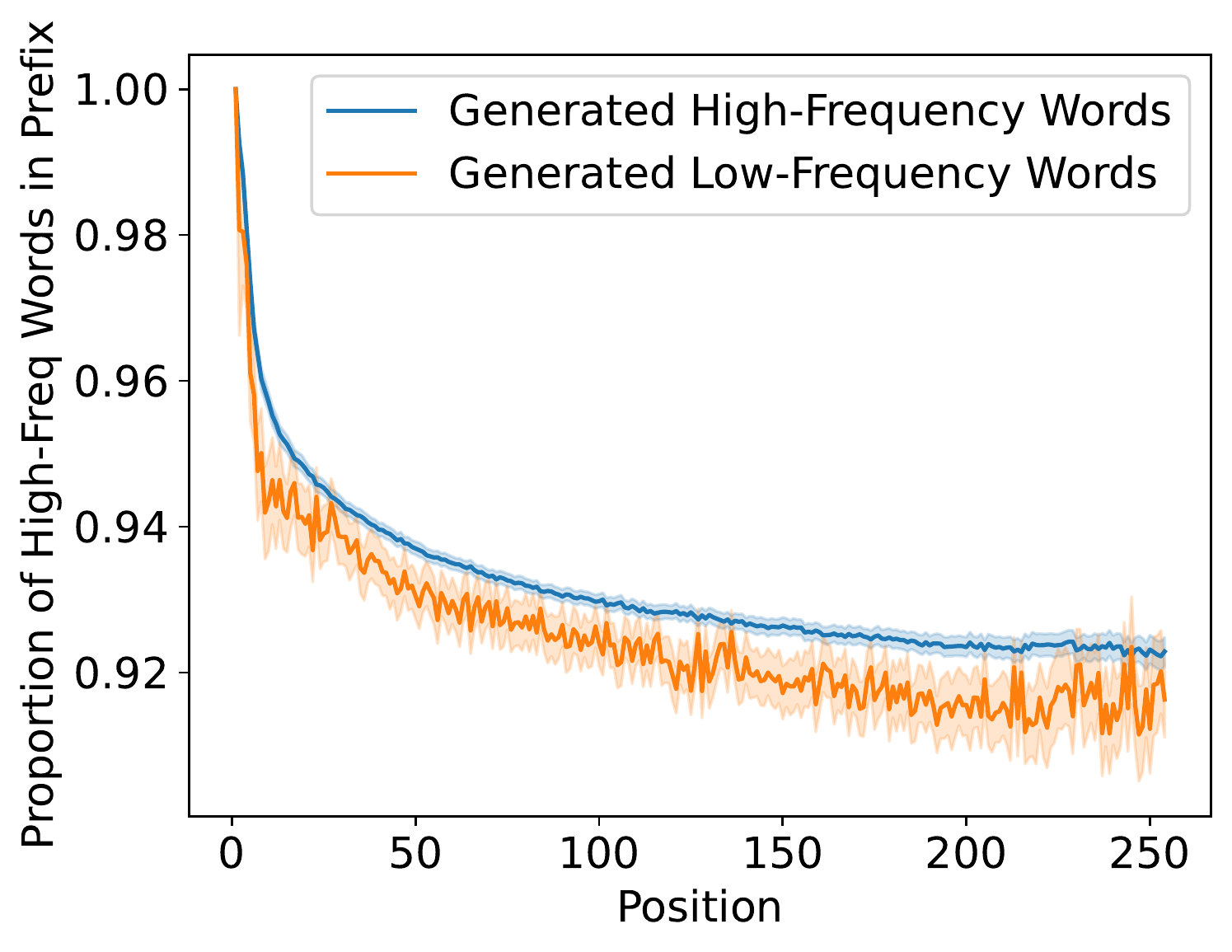}
\caption{The distance of embeddings in/between different frequency intervals on the Yelp Dataset, with the same setting in Figure~\ref{fig_propo}.} 
\label{fig_propo_yelp} 
\end{figure}

\begin{figure}[ht]
\center
\includegraphics[width=0.5 \textwidth]{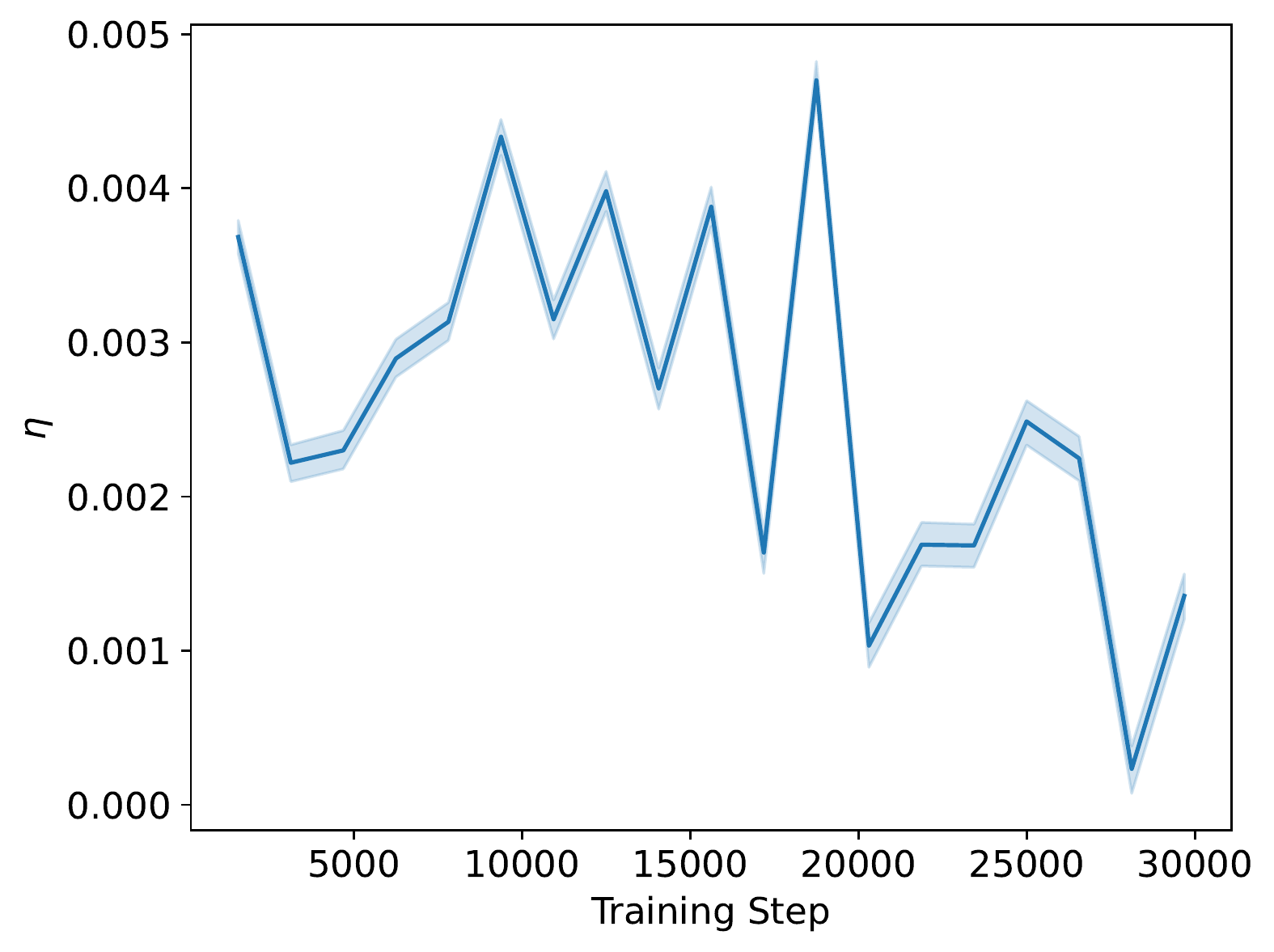}
\caption{The value of $\eta$ in Sub-claim 4 during training on the Yelp dataset, with the same setting in Figure~\ref{fig_sanity_roc}.} 
\label{fig_sanity_yelp} 
\end{figure}

\begin{figure}[tp]
\centering
\includegraphics[width=0.5\textwidth]{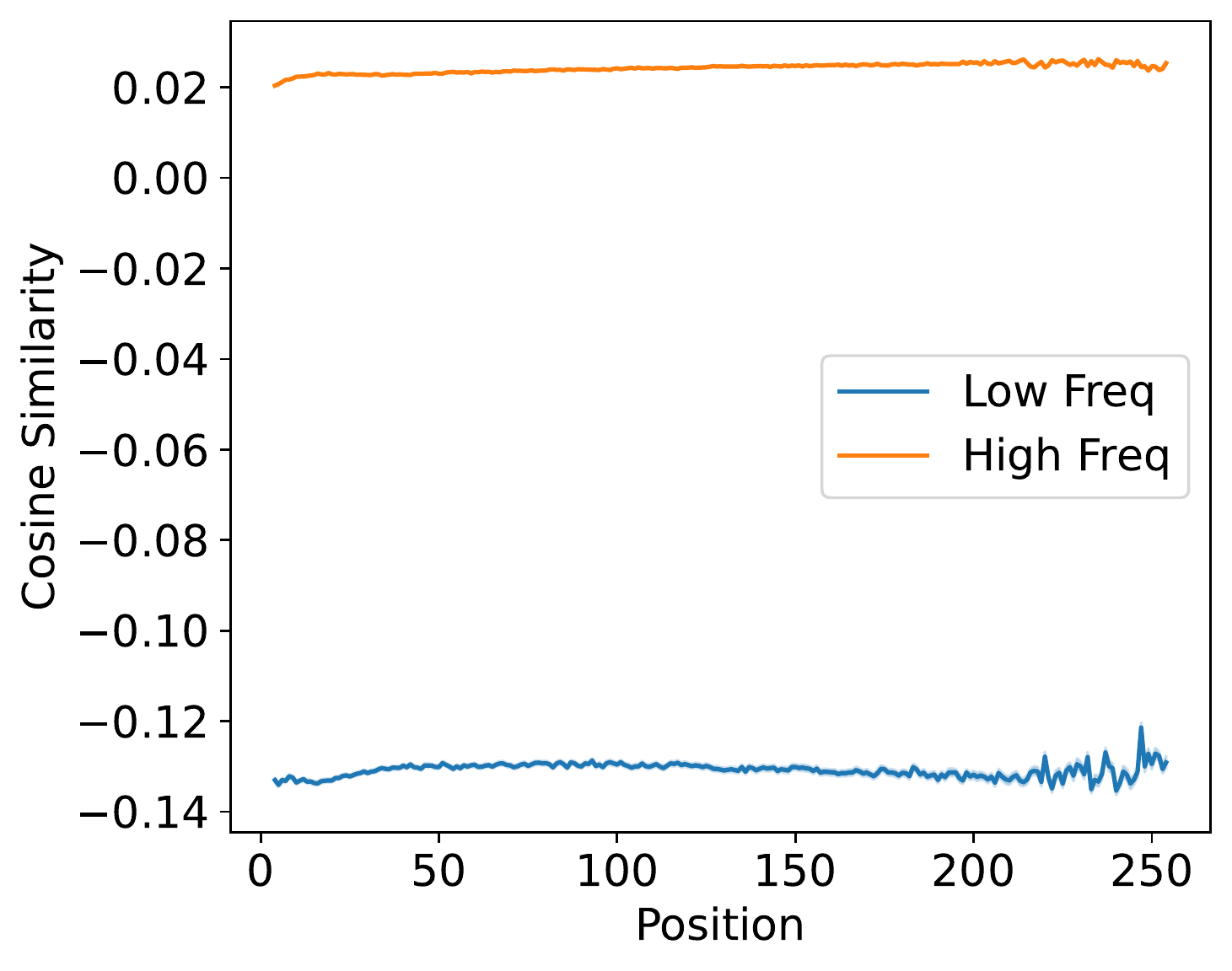}
\caption{The pairwise cosine similarity between the hidden state when generation and the high-frequency and low-frequency word embeddings in the Yelp dataset, with the same setting in Figure~\ref{cosdic_roc}.} 
\label{cosdic_yelp} 
\end{figure}

We conducted the same experiment in Section 2 on the Yelp Dataset. We correspondingly demonstrate the result of Figure ~\ref{fig_freq_roc}, ~\ref{fig_mat_roc}, ~\ref{fig_propo}, ~\ref{fig_sanity_roc} and ~\ref{cosdic_roc} in Figure ~\ref{fig_freq_yelp}, ~\ref{fig_mat_yelp}, ~\ref{fig_propo_yelp}, ~\ref{fig_sanity_yelp} and ~\ref{cosdic_yelp}. The results all indicate the same result as in ROC stories, demonstrating the generalizability of our theory across different datasets and across the conditional and unconditional generation tasks.

\section{Derivation and Proof}
\label{appc}

\subsection{The Demonstration for the Uniformly Positive Direction in Sub-claim 3}

\citet{RepDeg} proves their Theorem 2 by separately proving the necessary and sufficient part. We will correspondingly modify them to prove the same conclusion in the uniformly positive direction. 

In the necessary part, they begin by assuming the convex hull contains the origin. This assumption leads to the conclusion that there exists $\alpha_i$ and a vector $\boldsymbol{v}$ satisfying $\sum_i\alpha_i\langle \boldsymbol{h}_i, \boldsymbol{v}\rangle=0$, which contradicts the existence of the uniformly negative direction, \textit{i.e.}, $\forall i, \langle \boldsymbol{h}_i, \boldsymbol{v}\rangle<0 $ for the hidden states $\boldsymbol{h}_i$. Meanwhile, this inducted form also contradicts the existence of the uniformly positive direction $\forall i, \langle \boldsymbol{h}_i, \boldsymbol{v}\rangle>0 $. Thus, the necessary part of the uniformly positive direction could be easily verified also by contradiction.

For the sufficient part, the authors illustrate the existence of a hyperplane containing the origin and apart from the convex hull. A normal direction of this hyperplane would be the uniformly negative direction. For modification, another normal direction opposite from the uniformly negative direction could also meet the requirement for the uniformly positive direction.

Combining the necessary and sufficient parts, we can know there is a uniformly positive direction for all hidden states. As the hidden states of frequent tokens account for the largest proportion of all hidden states (Pareto principle), such a uniform direction tends to be closer to hidden states corresponding to the frequent tokens, successfully proving sub-claim 3.

\subsection{The Derivation of Sub-claim 4}
For some specific word with index $k$ in a sentence $x$ with length $m$, assuming that this word only occurs on position $m$, that is, only $x_m=k$. Let $p_i = p(i=k|\boldsymbol{h}_i)$, then we have 
\begin{equation}
\frac{\partial L_{NLL}}{\partial \boldsymbol{w}_k}=(1-p_m)\boldsymbol{h}_m+\sum_{i=1}^{m-1}p_i\boldsymbol{h}_i.
\end{equation}

Although we could yield the conclusion from this formula that the optimization direction of the word embedding agrees with the direction of hidden states (since $1-p_m$ and $p_i$ are all positive), the contributions of the hidden states are highly biased by $p_i$. In the early stage of training, $p_i \approx \frac1V$ for all $i$, thus the optimization direction will be mainly directed by $\boldsymbol{h}_m$. While in the late stage of training, $p_i \approx 0$ for $i \neq m$, and $p_{i, i \neq m} \ll p_m \approx \epsilon < 1$, the direction of $\boldsymbol{h}_m$ will also dominate the optimization. We could barely tell that the optimization direction is the uniformly positive direction from the final layer. Thus we resort to the previous layer.

For simplification, we ignore the nonlinear layer and let $\boldsymbol{h}_i=\sum_ja_{i,j}\hat{\boldsymbol{h}}_j$, where $\hat{\boldsymbol{h}}_j$ is the hidden state passed from the previous layer in Transformer, then we have:
\begin{equation}\begin{aligned} 
    \frac{\partial L_{NLL}}{\partial \boldsymbol{w}_k}&=\sum_{i=1}^{m-1}\left[a_{m,i}(1-p_m)-\sum_{j=i}^{m-1}a_j^ip_j\right]\hat{\boldsymbol{h}}_i
\\&+(1-p_m)a_{m,m}\hat{\boldsymbol{h}}_m.
\end{aligned}\end{equation}

We then consider the corner case when the attention distribution reaches the highest entropy, which leads to $a_{i,k}=\frac1i$. In this case, we have
\begin{equation}\begin{aligned} 
\frac{\partial L_{NLL}}{\partial \boldsymbol{w}_k}&=\sum_{i=1}^{m-1}(\frac{1-p_m}m-\sum_{j=i}^{m-1}\frac{p_j}j)\hat{\boldsymbol{h}}_i
\\&+\frac{1-p_m}m\hat{\boldsymbol{h}}_m\\
&=\sum_{i=1}^{m}(\frac1m-\sum_{j=i}^{m}\frac{p_j}j)\hat{\boldsymbol{h}}_i
\label{equ9} 
\end{aligned}\end{equation}

Thus, if $ \frac1m-\sum_{j=i}^{m}\frac{p_j}j > 0$ always stands, the optimization direction of the embeddings would approach the mixture of the hidden states in the preceding context. If a uniformly positive direction of the hidden states exists, i.e. $\hat{\boldsymbol{h}}_{t}\boldsymbol{v}^T> 0$ stands for all the hidden states, then we have 
\begin{equation}\begin{aligned} 
\frac{\partial L_{NLL}}{\partial \boldsymbol{w}_k}\boldsymbol{v}^T&=
\sum_{i=1}^{m}(\frac1m-\sum_{j=i}^{m}\frac{p_j}j)\hat{\boldsymbol{h}}_i\boldsymbol{v}^T> 0.
\end{aligned}\end{equation}

Thus, the direction $\boldsymbol{v}$ is also a uniformly positive direction for the optimization direction of the word embeddings. Moreover, the gradient contribution of $\hat{\boldsymbol{h}}_i$ is more scattered among the hidden states, and the tendency will be advanced when considering more previous layers. Therefore, we could conclude that the optimization will approach the uniformly positive direction under this condition.

\subsection{The Proof for \textbf{Theorem 1}}
We start our proof by the definition of the Rényi entropy
\begin{equation}\begin{aligned}
    \mathrm{H}_{\alpha}(\boldsymbol{a}_t)&=\frac1{1-\alpha}\log(\sum_ia_{t,i}^{\alpha})\\
    &=\frac1{1-\alpha}\log(\sum_i\frac{e^{\alpha \tilde{a}_{t, i}}}{Z^{\alpha}})\\
    &=\frac1{1-\alpha}\left[\log(\sum_i e^{\alpha \tilde{a}_{t, i}}) -\alpha \log Z\right]
\end{aligned}\end{equation}
Then we let $\alpha > 1$, then
\begin{equation}
\mathrm{H}_{\alpha}(\boldsymbol{a}_t)=\frac1{\alpha-1}[\alpha \log Z-\log(\sum_i e^{\alpha \tilde{a}_{t, i}})]
\end{equation}
According to the Jensen Inequality, 
 \begin{equation}\begin{aligned}
\log(\sum_i e^{\alpha \tilde{a}_{t, i}}) &= \log(\sum_i \frac{e^{\alpha \tilde{a}_{t, i}}}t) + \log t \\
&\geq \frac{\sum_i \log e^{\alpha \tilde{a}_{t, i}}}t + \log t\\
&=\frac{\alpha\sum_i\tilde{a}_{t, i}}t + \log t
\end{aligned}\end{equation}
 \begin{equation}\begin{aligned}
\therefore \mathrm{H}_{\alpha}(\boldsymbol{a}) &\leq \frac{1}{\alpha-1}\left[\alpha \log Z-\frac{\alpha}{t} \sum \tilde{a}_{t, i}-\log t\right] \\
&\leq \frac{1}{\alpha-1}\left[\alpha \log Z-\frac{\alpha}{t} \sum \tilde{a}_{t, i}\right]
\end{aligned}\end{equation}
While
\begin{equation}\begin{aligned} 
    \log Z&=\log \sum e^{\tilde{a}_{t, i}} \\
    &\leq \max \{\tilde{a}_{t, 1}, \ldots \tilde{a}_{t, n}\}+\log t\\
    & \leq \max \{ |\tilde{a}_{t, 1}|, \ldots|\tilde{a}_{t, n}| \} +\log t \\
    &=\Vert\boldsymbol{\tilde{a}}_t\Vert_\infty+\log t
\end{aligned}\end{equation}
\begin{equation}
\begin{aligned}
\therefore \mathrm{H}_\alpha(\boldsymbol{a}_t) & \leq \frac{1}{\alpha-1}\bigg[\alpha\Vert\boldsymbol{\tilde{a}}_t\Vert_\infty+\alpha \log t -\frac{\alpha}{t} \sum_{i} \tilde{a}_{t, i}-\log t\bigg] \\
& \leq \frac{\alpha}{\alpha-1}\left[\Vert\boldsymbol{\tilde{a}}_t\Vert_\infty-\frac{1}{t} \sum_{i} \tilde{a}_{t, i}\right]+ \log t \\
& \leq \frac{\alpha}{\alpha-1}\left[\Vert\boldsymbol{\tilde{a}}_t\Vert_1-\frac{1}{t} \sum_{i} \tilde{a}_{t, i}\right] + \log t
\end{aligned}
\end{equation}
Then we consider the term
 \begin{equation}\begin{aligned}
\Vert\boldsymbol{\tilde{a}}_t\Vert_1-\frac{1}{t} \sum_{i} \tilde{a}_{t,i}&=\sum_{i}\left[\left|\tilde{a}_{t,i}\right|-\frac{1}{t} \tilde{a}_{t,i}\right]\\
&\leq \sum_{i}\left[\left(1+\frac{1}{t}\right)\left|\tilde{a}_{t,i}\right|\right]
\end{aligned}
\end{equation}
\begin{equation}
\begin{aligned}
\therefore 
\mathrm{H}_\alpha(\boldsymbol{a}_t) &\leq\frac{\alpha\left(1+\frac{1}{t}\right)}{\alpha-1} \sum_{i}\left|\tilde{a}_{t, i}\right|+ \log t\\
&=\frac{\alpha\left(1+\frac{1}{t}\right)}{\alpha-1}\Vert\boldsymbol{\tilde{a}}_t\Vert_{1}+ \log t\\
&=\frac{\alpha(t+1)}{t(\alpha-1)}\Vert\boldsymbol{\tilde{a}}_t\Vert_{1}+ \log t \\
&\leq \frac{\alpha(t+1)}{t(\alpha-1)}\Vert\boldsymbol{\hat{a}}_t\Vert_{1}+ \log t + \frac{\alpha(t+1)}{\alpha-1} |C|
\end{aligned}\end{equation}
where $C$ is a large negative constant used in attention dropout. 

Discarding all the constants, minimizing the upper bound of $\mathrm{H}_\alpha(\boldsymbol{a}_t)$ above is equal to minimize
\begin{equation}
\frac{\alpha(t+1)}{t(\alpha-1)}\Vert\boldsymbol{\hat{a}}_t\Vert_{1}
\end{equation}

\subsection{The Proof for \textbf{Theorem 2}}

We use the Bayes theorem to formalize the posterior distribution as follows:

\begin{equation}
\begin{aligned}
&KL\left[q_{\theta}(\boldsymbol{\tilde{a}}_t)|| p(\boldsymbol{\tilde{a}}_t | x_t,y_t, c)\right]\\
&=\int q_{\theta}(\boldsymbol{\tilde{a}}_t) \log \frac{q_{\theta}(\boldsymbol{\tilde{a}}_t)}{p(\boldsymbol{\tilde{a}}_t | x_t,y_t, c)} \dif \boldsymbol{\tilde{a}}_t\\
&=\int q_{\theta}(\boldsymbol{\tilde{a}}_t) \log \frac{q_{\theta}(\boldsymbol{\tilde{a}}_t)}{p(\boldsymbol{\tilde{a}}_t | x_t,y_t, c)}\cdot \frac{p(\boldsymbol{\tilde{a}}_t)}{p(\boldsymbol{\tilde{a}}_t)} \dif \boldsymbol{\tilde{a}}_t\\
&=KL \left[q_{\theta}(\boldsymbol{\tilde{a}}_t) || p(\boldsymbol{\tilde{a}}_t)\right] + \int q_{\theta}(\boldsymbol{\tilde{a}}_t) \log \frac{p(\boldsymbol{\tilde{a}}_t)}{p(\boldsymbol{\tilde{a}}_t | x_t,y_t, c)} \dif \boldsymbol{\tilde{a}}_t\\
&=KL \left[q_{\theta}(\boldsymbol{\tilde{a}}_t) || p(\boldsymbol{\tilde{a}}_t)\right] - \int q_{\theta}(\boldsymbol{\tilde{a}}_t) \log \frac{p(\boldsymbol{\tilde{a}}_t | x_t,y_t, c)}{p(\boldsymbol{\tilde{a}}_t)} \dif \boldsymbol{\tilde{a}}_t\\
\end{aligned}
\end{equation}
\begin{equation}
\begin{aligned}
\because p(\boldsymbol{\tilde{a}}_t | x_t,y_t,c) &= \frac{p(\boldsymbol{\tilde{a}}_t)p(x_t,y_t, c|\boldsymbol{\tilde{a}}_t)}{p(x_t,y_t, c)}\\
\therefore \frac{p\left(\boldsymbol{\tilde{a}}_t|x_t, y_t, c\right)}{p\left(\boldsymbol{\tilde{a}}_t\right)}&=\frac{p\left(x_t, y_t, c|\boldsymbol{\tilde{a}}_t\right)}{p(x_t, y_t, c)}\\
&=\frac{p\left(y_t|x_t, c, \boldsymbol{\tilde{a}}_t\right)p\left(x_t, c | \boldsymbol{\tilde{a}}_t\right)}{p(y_t|x_t, c)p(x_t, c)}\\
&=\frac{p\left(y_t|x_t,c,\boldsymbol{\tilde{a}}_t\right)}{p(y_t|x_t, c)}
\end{aligned}\end{equation}
\begin{equation}
\begin{aligned}
\therefore&\quad KL\left[q_{\theta}(\boldsymbol{\tilde{a}}_t)|| p(\boldsymbol{\tilde{a}}_t | x_t,y_t,c)\right]\\
&=KL \left[q_{\theta}(\boldsymbol{\tilde{a}}_t) || p(\boldsymbol{\tilde{a}}_t)\right] - \int q_{\theta}(\boldsymbol{\tilde{a}}_t) \log \frac{p\left(y_t|x_t,c,\boldsymbol{\tilde{a}}_t\right)}{p(y_t|x_t,c)} \dif \boldsymbol{\tilde{a}}_t\\
&=KL \left[q_{\theta}(\boldsymbol{\tilde{a}}_t) || p(\boldsymbol{\tilde{a}}_t)\right] - E_{q_\theta(\boldsymbol{\tilde{a}}_t)} \left[ \log p\left(y_t|x_t, c,\boldsymbol{\tilde{a}}_t\right)\right]+\log p(y_t|x_t,c)
\end{aligned}\end{equation}
Thus, the process to minimize the KL Divergence $KL\left[q_{\theta}(\boldsymbol{\tilde{a}}_t)|| p(\boldsymbol{\tilde{a}}_t | x_t,y_t,c)\right]$ is equivalent to maximize
\begin{equation}
\begin{aligned}
&-KL\left[q_{\theta}(\boldsymbol{\tilde{a}}_t)|| p(\boldsymbol{\tilde{a}}_t | x_t,y_t,c)\right]\\
&=E_{q_\theta(\boldsymbol{\tilde{a}}_t)} \left[ \log p\left(y_t|x_t,c, \boldsymbol{\tilde{a}}_t\right)\right]-KL \left[q_{\theta}(\boldsymbol{\tilde{a}}_t) || p(\boldsymbol{\tilde{a}}_t)\right]-\log p(y_t|x_t,c)\\
&\geq E_{q_\theta(\boldsymbol{\tilde{a}}_t)} \left[ \log p\left(y_t|x_t, c,\boldsymbol{\tilde{a}}_t\right)\right]-KL \left[q_{\theta}(\boldsymbol{\tilde{a}}_t) || p(\boldsymbol{\tilde{a}}_t)\right]
\end{aligned}\end{equation}
We denote 
\begin{equation}
    q_{\theta}(\boldsymbol{\tilde{a}}_t)=p\delta(\boldsymbol{\tilde{a}}_t-C) + (1-p)\delta(\boldsymbol{\tilde{a}}_t-\boldsymbol{\hat{a}}_{t})\\
\end{equation}
where $C$ is a large negative number, and p is the probability of the attention dropout.
We define $p(\boldsymbol{\tilde{a}}_t) \propto 1 + e^{-\gamma\Vert\boldsymbol{\tilde{a}}_t\Vert}$, then we have
\begin{equation}
\begin{aligned}
KL \left[q_{\theta}(\boldsymbol{\tilde{a}}_t) || p(\boldsymbol{\tilde{a}}_t)\right] &= \int q_{\theta}(\boldsymbol{\tilde{a}}_t) \log \frac{q_{\theta}(\boldsymbol{\tilde{a}}_t)}{p(\boldsymbol{\tilde{a}}_t)} \dif \boldsymbol{\tilde{a}}_t\\
&=\int q_{\theta}(\boldsymbol{\tilde{a}}_t) \log \frac{q_{\theta}(\boldsymbol{\tilde{a}}_t)}{1+e^{-\gamma\Vert\boldsymbol{\tilde{a}}_t\Vert}} \dif \boldsymbol{\tilde{a}}_t + \log Z\\
&=p \log p + (1-p)\log \frac{1-p}{1+e^{-\gamma\Vert\boldsymbol{\hat{a}}_t\Vert}} + \log Z\\
&=p \log p + (1-p)\log(1-p)+(1-p)\log(1+e^{-\gamma\Vert\boldsymbol{\hat{a}}_t\Vert})+\log Z\\
&=-\mathrm{H}(p)+(1-p)\log(1+e^{-\gamma\Vert\boldsymbol{\hat{a}}_t\Vert}) + \log Z
\end{aligned}\end{equation}
According to the Jensen Inequality, we have
\begin{equation}
\begin{aligned}
\frac{e^x+e^{-x}}2 &\geq e^{\frac{x}2}\\
\therefore 1+e^x &\geq 2e^{\frac{x}2}\\
\therefore log(1+e^x) &\geq \log 2 + \frac{x}2
\end{aligned}
\end{equation}
\begin{equation}
\begin{aligned}
\therefore KL \left[q_{\theta}(\boldsymbol{\tilde{a}}_t) || p(\boldsymbol{\tilde{a}}_t)\right]  &\leq -\mathrm{H}(p)-(1-p)\left[log2+\frac{-\gamma\Vert\boldsymbol{\hat{a}}_t\Vert}2\right] + \log Z\\
&=-\mathrm{H}(p)+\frac{\gamma(1-p)}2\Vert\boldsymbol{\hat{a}}_t\Vert-(1-p)\log2 + \log Z
\end{aligned}
\end{equation}
Discard all the constants, and we have 
\begin{equation}
-\mathrm{H}(p)+\frac{\gamma(1-p)}2\Vert\boldsymbol{\hat{a}}_t\Vert
\end{equation}

\section{Additional Results}
\label{appd}
\subsection{Result with \textsc{Care}-O}

In Section 3, we mentioned a variant of our method only with the $\mathcal{L}_R $ term remaining the original dropout pattern. In this section, we call this variant \textsc{Care}-O, and examine its performance on ROC, Yelp, and ParaSCI. The results are listed in Table ~\ref{table_cond_gene_o} and Table ~\ref{res_unconditional_o}. As we can see, although it could achieve the best in some metrics like CND in ParaSCI, its overall quality is worse than GPT-2 only with a small margin on diversity improvement on ROC and ParaSCI. It also performs very poorly on Yelp. This experimental result agrees with our theoretical induction of the necessity of dropout modification.

\begin{table}[h]
\centering
\scalebox{0.8}{
\begin{tabular}{c|cccccccc|ccc}
\toprule
\multirow{2}{*}{Model} & \multicolumn{8}{c|}{Quality}                          &  \multicolumn{3}{c}{Diversity} \\ \cline{2-12}
                       & R-2$\uparrow$   & R-3$\uparrow$    & R-L$\uparrow$   & R-W$\uparrow$  & B-2$\uparrow$  & B-4$\uparrow$     & BS$\uparrow$     & CND$\downarrow$    &  Dist$\uparrow$      & JS$\downarrow$       & SB$\downarrow$      \\ \midrule
\multicolumn{11}{c}{Dataset: ParaSCI}\\
\midrule
GPT-2    & 41.48 & 32.65 & 54.97 & 35.28 & \textbf{41.64} & \underline{26.58} & 90.81 & 1.688 & 60.34 & 0.0726 & 17.52 \\
\hline
Pattern  & 41.46 & 32.58 & 55.06 & 35.34 & 41.31 & 26.43 & \underline{90.89} & 1.652 & 60.65 & 0.0683 & 16.86 \\
Entmax   & 38.44 & 29.78 & 52.28 & 33.50 & 38.06 & 23.60 & 90.35 & 1.769 & 59.30 & 0.0707 & 18.10 \\
$l_0$-Drop   & 37.26 & 28.64 & 50.99 & 32.53 & 37.96 & 23.47 & 90.05 & 1.700 & \underline{60.94} & \underline{0.0591} & \underline{16.18} \\
$\mathcal{L}_{\mathrm{A}}$-Tuning & 40.71 & 32.05 & 54.20 & 34.89 & 39.95 & 25.48 & 90.70 & 1.632 & 60.67 & 0.0622 & 16.28 \\ \hline
\textsc{Care}-O    & 41.06 & 32.59 & 54.16 & 35.09 & 38.30 & 24.28 & 90.61 & \textbf{1.626} & 59.21 & 0.0685 & 17.28 \\
\textsc{Care}    & \textbf{42.49} & \textbf{33.74} & \underline{55.69} & \textbf{35.94} & 41.59 & \textbf{26.75} & 90.80 & \underline{1.631} & \textbf{61.04} & \textbf{0.0566} & \textbf{16.05} \\
\textsc{Care}-A & \underline{41.95} & \underline{33.01} & \textbf{55.79} & \underline{35.76} & \underline{41.63} & 26.56 & \textbf{91.04} & 1.637 & 60.60 & 0.0596 & 17.04\\
\midrule
\multicolumn{11}{c}{Dataset: ROCStories}\\
\midrule
GPT-2    & 6.700 & 1.321 & 23.17 & \underline{11.19} & 5.817 & 0.266 & 83.53 & 8.856 & 25.72 & 0.6625 & 62.36 \\
\hline
Pattern  & 6.690 & 1.301 & 23.33 & 11.09 & 6.207 & 0.273 & \underline{83.64} & 9.006 & 27.20 & 0.6280 & 61.16 \\
Entmax   & 6.011 & 1.011 & 22.42 & 10.80 & 5.374 & 0.150 & 82.91 & 9.521 & 25.03 & 0.7317 & 62.57 \\
$l_0$-Drop   & 5.850 & 1.065 & 22.11 & 10.77 & 4.896 & 0.177 & 83.17 & \textbf{8.489} & 24.82 & 0.6207 & 62.72 \\
$\mathcal{L}_{\mathrm{A}}$-Tuning & 6.367 & 1.214 & 22.94 & 10.96 & 5.809 & 0.226 & 83.37 & 8.904 & \underline{27.24} & \underline{0.5905} & 61.06 \\ \hline
\textsc{Care}-O     & 5.984 & 1.068 & 22.52& 10.78 & 5.401 & 0.199 & 83.41 & \underline{8.603} & 26.78 & 0.5915 & \underline{61.0}5 \\
\textsc{Care}     & \underline{6.905} & \textbf{1.399} & \underline{23.54} & 11.08 & \textbf{6.829} & \textbf{0.326} & \textbf{83.74} & 9.735 & \textbf{27.77} & 0.6328 & \textbf{60.41} \\
\textsc{Care}-A   & \textbf{7.002} & \underline{1.332} & \textbf{23.80} & \textbf{11.32} & \underline{6.489} & \underline{0.274} & 83.63 & 8.742 & 25.89 & \textbf{0.5708} & 61.51\\
\bottomrule
\end{tabular}
}
\caption{The evaluation result of conditional generation with \textsc{Care}-O}
\label{table_cond_gene_o}
\end{table}

\begin{table}[ht]
\centering
\begin{tabular}{c|cccc|ccc}
\toprule
\multirow{2}{*}{Model} & \multicolumn{4}{c|}{Quality}                          &  \multicolumn{3}{c}{Diversity} \\ \cline{2-8}
& B-2$\uparrow$  & B-4$\uparrow$    & CND$\downarrow$   & MV$\uparrow$ &  Dist$\uparrow$     & JS$\downarrow$       & SB$\downarrow$      \\ \midrule
GPT-2    &76.68 & 36.46 & \underline{0.886} & \underline{0.931} & 27.62 & 0.187 & 47.76 \\
Pattern  &76.96 & \underline{38.09} & 0.924 & 0.859 & 27.34 & 0.203 & 49.31 \\
Entmax   &77.03& 32.18 & 0.951 & 0.920 & 23.99 & 0.222 & 49.18 \\
$l_0$-Drop   &76.71 & 34.92 & \textbf{0.866} & 0.919 & 26.33 & 0.194 & 47.38 \\
$\mathcal{L}_{\mathrm{A}}$-Tuning &\underline{77.35} & \textbf{38.66} & 0.910 & 0.835 & 26.26 & 0.211 & 50.82 \\ \hline

\textsc{Care}-O     &76.31 & 26.87 & 1.262 & 0.5128 & \textbf{17.10} & \underline{0.224} & \textbf{50.53} \\
\textsc{Care}     &73.84 & 33.61 & 1.033 & 0.863 & \textbf{30.84} & \underline{0.178} & \textbf{43.60} \\
\textsc{Care}-A & \textbf{77.53} & 35.69 & 0.933 & \textbf{0.974} & \underline{28.44} & \textbf{0.169} & \underline{44.65}\\
\bottomrule
\end{tabular}
\caption{Evaluation results for unconditional generation on the Yelp dataset with \textsc{Care}-O. MV stands for MAUVE.}
\label{res_unconditional_o}
\end{table}
\subsection{Result on MP}

We also examine our model on MP dataset. The result is in~\ref{table_mp} here due to the space limit. As we can see, our model achieves the best diversity and novelty with a little sacrifice of the generated quality, still demonstrating its ability to enhance the novelty and the diversity of the generated text.

\begin{table}[htbp]
\centering
\scalebox{0.8}{
\begin{tabular}{c|cccccccc|ccc}
\toprule
\multirow{2}{*}{Model} & \multicolumn{8}{c|}{Quality}                          &  \multicolumn{3}{c}{Diversity} \\ \cline{2-12}
                       & R-2$\uparrow$   & R-3$\uparrow$    & R-L$\uparrow$   & R-W$\uparrow$  & B-2$\uparrow$  & B-4$\uparrow$     & BS$\uparrow$     & CND$\downarrow$    &  Dist$\uparrow$      & JS$\downarrow$       & SB$\downarrow$      \\ \midrule
\midrule
GPT-2    & \underline{20.18} & \underline{12.76} & \underline{34.14} & \underline{24.47} & \underline{18.65} & \underline{9.817} & \underline{88.17} & 0.468 & 62.29 & 0.0159 & 17.17 \\
Pattern  & \textbf{20.34} & \textbf{12.82} & \textbf{34.27} & \textbf{24.59} & \textbf{18.77} & \textbf{9.850} & \textbf{88.23} & 0.470 & 62.26 & 0.0157 & 17.24 \\
$l_0$-Drop  & 19.65 & 12.00 & 33.78 & 24.14 & 18.21 & 8.838 & 88.14 & 0.480 & 62.38 & 0.0163 & 17.78 \\
\hline
\textsc{Care}-O    & 19.84 & 12.60 & 33.59 & 24.14 & 18.29 & 9.641 & 88.04 & \textbf{0.383} & \textbf{62.77} & \underline{0.0138} & 16.98 \\
\textsc{Care}    &19.49 & 12.31 & 33.24 & 23.87 & 17.98 & 9.625 & 88.02 & \underline{0.403} & \underline{62.76} & 0.0140 & \underline{16.90} \\
\textsc{Care}-A &18.84 & 11.73 & 32.61 & 23.40 & 17.14 & 8.703 & 87.79 & 0.407 & 62.73 & \textbf{0.0129} & \textbf{15.88} \\
\bottomrule
\end{tabular}
}
\caption{The evaluation result on headline generation.}
\label{table_mp}
\end{table}

\begin{table}[htbp]
\centering
\begin{tabular}{c|c}
\toprule
Model&Rep $\downarrow$      \\ \midrule
GPT-2    &5.598 \\
Pattern  &\textbf{4.571} \\
Entmax   &5.170 \\
$l_0$-Drop   &5.479 \\
$\mathcal{L}_{\mathrm{A}}$-Tuning &6.005 \\ \hline

\textsc{Care}     &\underline{4.742} \\
\textsc{Care}-A & 5.479\\
\bottomrule
\end{tabular}
\caption{Evaluation results of Rep on ROCStories Dataset.}
\label{res_rep_roc}
\end{table}

\subsection{Result on the repetition }
Although our model focus on improving inter-instance diversity instead of alleviating the intra-sentence repetition problem, we still add the result of Rep to examine the ability of our model to relieve the repetition. First used in \citep{DBLP:conf/iclr/WelleckKRDCW20}, Rep measures the percentage of repetitive generated n-grams. We implement this by ourselves and present the result of the geometric mean of Rep1-4 in ROCStories Dataset in Table \ref{res_rep_roc}.

As demonstrated, the method with sparse attention can achieve a better score of Rep, proving that sparse attention can enhance inter-instance diversity and soothe intra-sentence degeneration. Among them, our \textsc{Care} and Pattern achieve significantly superior scores compared to others, demonstrating the \textsc{Care}'s another efficacy in the reduction of repetition.

\clearpage
\section{Python Implementation of \textsc{Care}}
\label{appe}
\label{app_code}
We list the code of \textsc{Care} as follows. This code is modified from the Huggingface Transformers Library \citep{wolf-etal-2020-transformers}. Apart from the necessary function name and class name for location, the newly added code contains 19 added or modified lines highlighted in blue.

\begin{figure}[htbp]
\centering
\includegraphics[trim={0 15cm 0 0},clip,width=1.0\textwidth]{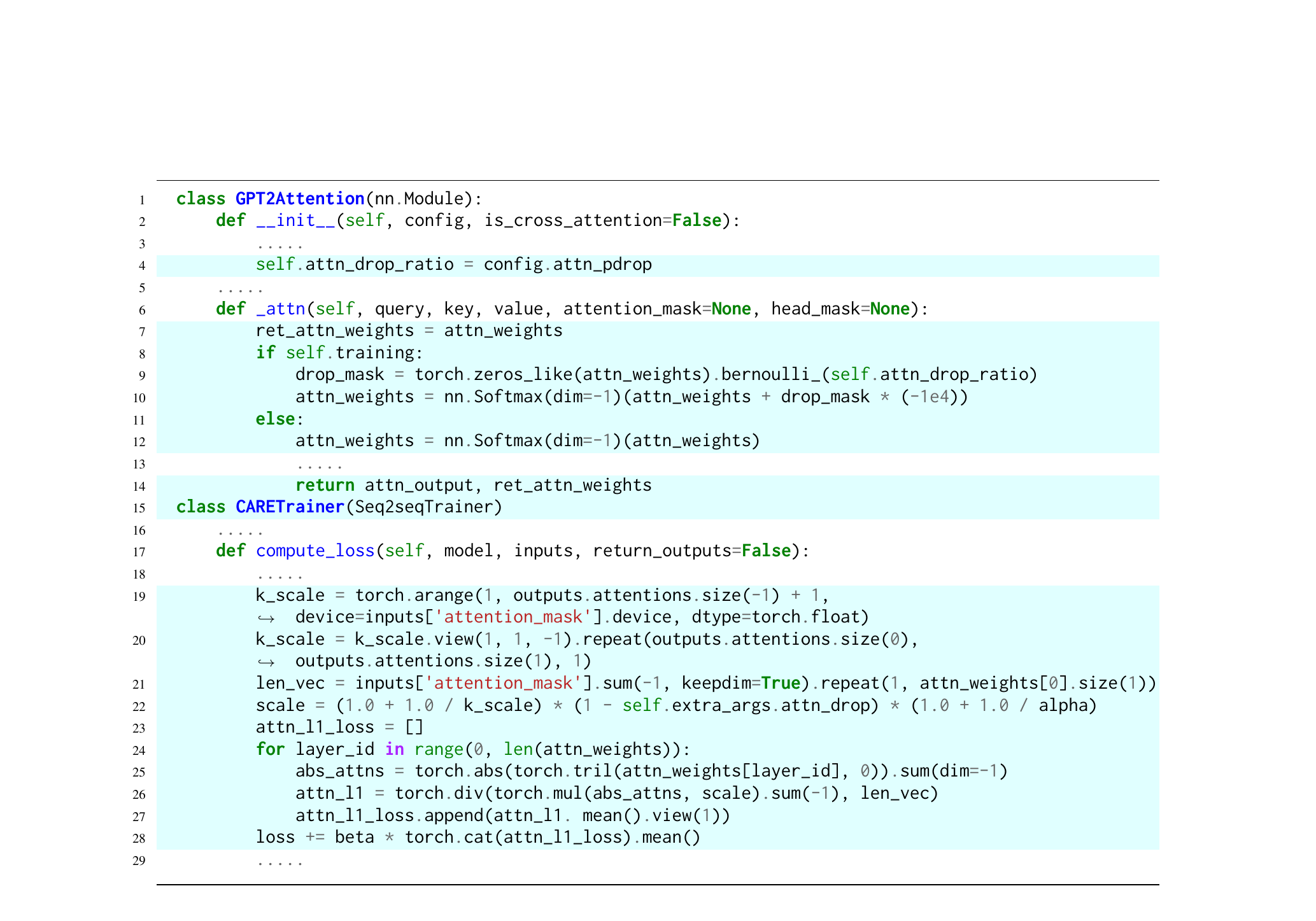}
\end{figure}

\end{document}